\documentclass[a4paper,fleqn]{cas-dc}

\usepackage{graphicx}
\usepackage[authoryear,longnamesfirst]{natbib}
\usepackage{textcomp}
\usepackage{amsmath}
\usepackage{amsthm}
\newtheorem{proposition}{Proposition}
\usepackage{subcaption}
\usepackage{hyperref}
\usepackage{amsfonts}
\usepackage{placeins}
\usepackage{multirow}
\usepackage{booktabs}
\usepackage{tabularx}
\usepackage{tikz}
\usetikzlibrary{positioning, arrows.meta, calc}

\usepackage{amsmath,amsfonts,bm,mathtools,xargs}
\usepackage{scalerel,stackengine}

\def\eqref#1{equation~\ref{#1}}

\def\1{\bm{1}}

\def\rvepsilon{{\boldsymbol{\epsilon}}}

\def\rvmu{{\boldsymbol{\mu}}}
\def\rvsigma{{\boldsymbol{\sigma}}}

\def\rvu{{\mathbf{i}}}

\def\rvu{{\mathbf{u}}}

\def\rvw{{\mathbf{w}}}
\def\rvx{{\mathbf{x}}}

\def\rvz{{\mathbf{z}}}

\def\rmI{{\mathbf{I}}}

\def\rmX{{\mathbf{X}}}
\def\rmY{{\mathbf{Y}}}

\def\vtheta{{\bm{\theta}}}
\def\vphi{{\bm{\phi}}}

\def\mD{{\bm{D}}}

\def\mX{{\bm{X}}}

\def\mSigma{{\bm{\Sigma}}}

\DeclareMathAlphabet{\mathsfit}{\encodingdefault}{\sfdefault}{m}{sl}
\SetMathAlphabet{\mathsfit}{bold}{\encodingdefault}{\sfdefault}{bx}{n}

\newcommand{\kld}{D_\mathrm{KL}}
\DeclarePairedDelimiterXPP\kl[2]{\kld}(){}{#1\;\delimsize\|\;#2}

\newcommandx{\Dbig}[4][3,4,usedefault]{D^{#4}_{\mathrm{#3}}\big(#1\;\|\;#2\big)}
\newcommandx{\DBig}[4][3,4,usedefault]{D^{#4}_{\mathrm{#3}}\Big(#1\;\big\|\;#2\Big)}
\newcommandx{\Dbigg}[4][3,4,usedefault]{D^{#4}_{\mathrm{#3}}\bigg(#1\;\Big\|\;#2\bigg)}
\newcommandx{\DBigg}[4][3,4,usedefault]{D^{#4}_{\mathrm{#3}}\Bigg(#1\;\bigg\|\;#2\Bigg)}

\newcommand{\Var}{\mathrm{Var}}

\newcommandx{\ELBO}[2][1,2,usedefault]{\mathcal{L}_{#1}(\vtheta, \vphi; \rvx #2)}

\makeatletter
\def\biggg{\bBigg@{3.5}}
\def\Biggg{\bBigg@{4}}
\makeatother

\ExplSyntaxOn
\cs_gset:Npn \__first_footerline:
{
  \group_begin:
  \small
  \sffamily
  \ifnum\theblind>0\relax
  \else
    \__short_authors:
  \fi
  \group_end:
}
\ExplSyntaxOff

\def\BibTeX{{\rm B\kern-.05em{\sc i\kern-.025em b}\kern-.08em
    T\kern-.1667em\lower.7ex\hbox{E}\kern-.125emX}}

\makeatletter
\def\printorcid#1{}
\makeatother

\def\tsc#1{\csdef{#1}{\textsc{\lowercase{#1}}\xspace}}
\tsc{WGM}
\tsc{QE}

\begin{document}
\let\WriteBookmarks\relax
\def\floatpagepagefraction{1}
\def\textpagefraction{.001}

\shortauthors{P. Clapham et~al.}
\shorttitle{Unifying Latent Collapse Metrics}

\title[mode = title]{Unifying Latent Collapse Metrics: What They Measure and What They Miss}

\author[aff1]{Peter Clapham}
\cormark[1]
\ead{pclapham129@gmail.com}
\author[aff2]{Lisa Bonheme}
\author[aff1]{Marek Grzes}

\affiliation[aff1]{
  organization={School of Computing, University of Kent},
  city={Canterbury},
  country={United Kingdom}
}

\affiliation[aff2]{
  organization={International Agency for Research on Cancer (IARC/WHO)},
  addressline={Genomic Epidemiology Branch},
  city={Lyon},
  country={France}
}

\cortext[1]{Corresponding author}

\begin{abstract}
    Latent-variable models often suffer \emph{collapse}: some latent dimensions stop encoding anything
    about the data. The field detects this with several disconnected tests, including a variance threshold
    frozen in 2016, per-dimension KL criteria, encoder conditions specific to the standard VAE, and
    mutual-information estimates. We place these criteria in a common account centred on the entropy of
    the mean representation: the KL mean term, AU variance, and entropy are nested predictive-family views,
    while sampled-code mutual information is bracketed by the corresponding entropy-power and variance
    quantities. Across 180 trained models spanning four architectures the calibrated entropy criterion and
    the frozen AU rule agree, differing on fewer than $1\%$ of dimensions.
 
    The same analysis exposes what these statistics share as a blind spot: they measure how much a
    dimension varies, not what its variation \emph{carries}. After correcting for the difference in scale we found that
    collapsed dimensions still decode a number of generative factors including the class label. Any model given more latent dimensions than the data require retains
    such content; identifiable VAEs retain it most strongly, their conditional prior leaving the label
    readable in nearly every collapsed dimension, where a capacity-matched ordinary VAE's collapsed
    dimensions sit at chance. Collapse suppresses amplitude, not content. Deciding that a dimension
    has collapsed is a solved problem; deciding what it still contains lies outside the vocabulary
    of every statistic in the literature.
\end{abstract}


\begin{keywords}
variational autoencoders \sep posterior collapse \sep information collapse \sep entropy \sep mutual information \sep active units \sep identifiability
\end{keywords}

\maketitle

\section{Introduction}
\label{sec:int}

Variational Autoencoders (VAEs) are widely used for unsupervised representation learning
\citep{Kingma2014AutoEncodingVB}. They aim to learn a low-dimensional yet informative representation of
the data, and for that representation to be useful, whether for generation or for downstream tasks, it
should capture as much relevant variation as possible.

\par

A recurring problem is \emph{collapse}: some latent dimensions stop encoding information about the data.
When every dimension collapses, the model suffers \emph{posterior collapse} \citep{Lucas2019UnderstandingPC},
degrading both generative quality and latent interpretability \citep{Bowman2016GeneratingSF}. More often
collapse is \emph{selective}: only a subset of dimensions collapse while the rest remain informative, defining the
\emph{polarised regime} \citep{dai2018diagnosing,rolinek2019VAEPCA}. In the polarised regime the
representation is described as splitting into three subsets \citep{Bonheme2021BeMA}: \emph{active} dimensions, which
encode the variation most useful for reconstruction; \emph{passive} dimensions, which have collapsed and
encode little or nothing, typically because the model adheres too tightly to the prior
\citep{Lucas2019UnderstandingPC,Wang2022PosteriorCO} or has more latent capacity than the data require
\citep{dai2018diagnosing,rolinek2019VAEPCA}; and \emph{mixed} dimensions, which behave as active for some
inputs and passive for others.

\par

The difficulty is not that the field lacks collapse criteria, but that it has several, disconnected ones.
A widely used criterion is the \emph{active-units} (AU) count of \cite{Burda2016IWAE}: a dimension
is active if the variance of its mean representation across the data exceeds a fixed constant,
$A_u = \mathrm{Cov}_{\rvx}(\mathbb{E}_{u\sim q(u|\rvx)}[u]) > 10^{-2}$. The metric is a
commonly reported column in the posterior-collapse literature
\citep{He2019Lagging,Norouzi2020Exemplar,Razavi2019Delta,Menon2022Forget}; yet the threshold is
arbitrary, and therefore can vary between papers (e.g.\ $2\times10^{-2}$ in \citealp{Yeung2017EpitomicVA}). More significantly, beyond its intuition it comes with no proof of its validity. Alongside AU sit criteria that threshold the per-dimension KL divergence against a Gaussian prior
\citep{Lucas2019UnderstandingPC, Bowman2016GeneratingSF}, encoder-statistic conditions specific to the standard VAE
\citep{Bonheme2021BeMA,rolinek2019VAEPCA}, and mutual-information estimates built from the rate term
\citep{Hoffman2016ELBOSurgery,Menon2022Forget}. These statistics agree on standard trained models, but no
theory connects them. Carrying any of them to another model class has meant introducing a new, equally
specific definition. For identifiable VAEs, the existing criterion was restated in terms of the conditional prior
\citep{bonheme2023the}. A deterministic autoencoder, which has no prior and no KL term, falls outside the
KL- and MI-based criteria entirely, and \cite{Lucas2019UnderstandingPC} note that despite the volume of
work on posterior collapse it has not been measured in a consistent way, or even consistently defined.

\par

We argue that these apparently distinct criteria admit a common account centred on the entropy of the
mean representation. A dimension is collapsed when its mean representation $\rvmu_i$ is approximately
constant across the data and therefore exhibits negligible data-dependent variation. We quantify this
variation using the entropy $\mathrm{H}(\rvmu_i)$. For each model, collapsed dimensions form a
low-entropy floor cluster, separated from the informative dimensions by a sharp fall-off. We calibrate
the collapse threshold at this separation and classify dimensions below it as collapsed. A directional
entropy--variance bound connects this statistic to the AU variance criterion; the KL optimum recovers
Bonheme's passive condition; and an entropy-power sandwich links the same mean-side quantities to
rate-based mutual information on the sampled code. The entropy statistic itself is defined directly on
the mean, makes no reference to a prior or to the variational structure, and therefore applies unchanged
to variational and deterministic models alike.

\par

\noindent We make three contributions:
\begin{itemize}
    \item \textbf{Unification.} We show that the entropy of the mean provides a common account of
    existing collapse criteria. The KL mean term, AU variance, and entropy are the same predictive
    quantity under nested family restrictions; together with the KL optimum and an entropy-power
    bound, this connects them to Bonheme's passive condition and rate-based mutual information
    (Section~\ref{sec:theory}; Figure~\ref{fig:bridge}).

    \item \textbf{Validation and calibration.} Across 180 trained models, the calibrated entropy
    criterion and the frozen AU threshold agree on more than $99\%$ of dimensions. The reason is
    structural rather than numerical: collapsed and non-collapsed dimensions are separated by a
    fall-off spanning $10^{3}$--$10^{4}$ standard deviations of the measured estimator floor.
    Calibration also reveals a distinction hidden by binary collapse decisions: $\beta$-VAE passive
    clusters lie at the estimator floor, whereas iVAE passive clusters lie systematically above it
    (Sections~\ref{sec:floor}, \ref{sec:emp_gen}, and~\ref{sec:emp_agree}).

    \item \textbf{The limitation of collapse metrics.} All of these criteria measure the amplitude of
    a dimension's variation, not what that variation carries. After controlling amplitude and
    cardinality, collapsed dimensions retain decodable label and generative-factor information.
    Surplus latent capacity is sufficient for this retention; the identifiable VAE's conditional
    prior additionally distributes the auxiliary label across collapsed dimensions at high
    per-dimension density (Section~\ref{sec:emp_content}).
\end{itemize}

\section{Preliminaries}
\label{sec:prelim}

\subsection{Variational Autoencoders}
\label{sec:vae}

VAEs \citep{Kingma2014AutoEncodingVB} learn latent representations by mapping
each input to a distribution in latent space rather than to a single point. The encoder typically
outputs the parameters of a diagonal Gaussian,
\begin{equation}
    q_\phi(\rvz|\rvx) = \mathcal{N}(\rvmu, \mSigma), \qquad \mSigma = \text{diag}(\rvsigma^2),
    \label{eqn:latent}
\end{equation}
and samples are drawn using the reparameterisation trick,
\begin{equation}
\rvz = \rvmu + \rvsigma \odot \rvepsilon, \qquad \rvepsilon \sim \mathcal{N}(0, \rmI),
\end{equation}
where $\odot$ is the Hadamard product. Training maximises the Evidence Lower Bound (ELBO),
\begin{equation}
    \mathcal{L}_{\beta\text{-VAE}} = \mathbb{E}_{q_\phi(\rvz|\rvx)} \big[ \log p_\theta(\rvx|\rvz) \big]
    - \beta D_{\text{KL}} \big( q_\phi(\rvz|\rvx) \,||\, p(\rvz) \big),
    \label{eqn:elbo}
\end{equation}
with the prior $p(\rvz)$ usually a standard Gaussian.

\subsection{Collapse and the Polarised Regime}
\label{sec:polarised_regime}

Collapse is usually described as disappearance of information from a latent dimension. It is useful to separate two cases.
\emph{Posterior collapse} is total: every dimension collapses toward the prior and the latent code becomes
uninformative. The \emph{polarised regime} is selective: only a subset collapse, and the latent space
partitions into \emph{active} dimensions, which encode the variation most informative for reconstruction
(maximising the log-likelihood term of Eq.~\ref{eqn:elbo}); \emph{passive} (collapsed) dimensions, which
encode little, having collapsed toward the prior (minimising the KL term); and \emph{mixed} dimensions,
active for some inputs and passive for others. A useful intuition treats the three as switches: active
always ``on'', passive always ``off'', mixed switching with the input. Our criterion targets the passive
switch; a central message of this paper, however, is that ``off'' is a statement about
\emph{amplitude}, Section~\ref{sec:emp_content} shows that a collapsed dimension can still carry content.

\par

Two conditions recur in accounts of how a polarised regime arises. The first is surplus capacity: when the
latent space offers more dimensions than the data require, the unneeded ones are free to collapse
\citep{dai2018diagnosing,rolinek2019VAEPCA}. The second is regularisation pressure: $\beta>1$ strengthens it, so the model activates only the most
informative dimensions and collapses the rest \citep{Higgins2017betaVAELB,rolinek2019VAEPCA}; but that
pressure need not come from a prior at all, as the deterministic models of Section~\ref{sec:setup} show.
Ideally the collapsed dimensions capture noise, but under heavy regularisation
genuine factors can also be suppressed.

\par

The most explicit structural definition of the polarised regime is that of \cite{Bonheme2021BeMA}, stated in terms of the mean
($\rvmu$) and variance ($\rvsigma$) representations. Writing $\bar{\rvmu}$ for the data-average of
$\rvmu$ and $\bar{\rvsigma}$ for the data-average of
$\rvsigma$:
\begin{itemize}
\label{def:bon}
    \item \textbf{Active}: $\bar{\rvsigma} \ll 1$;
    \item \textbf{Passive}: $\bar{\rvsigma} \approx 1$, $\Var(\rvsigma) \ll 1$, $|\bar{\rvmu}| \ll 1$,
    $\Var(\rvmu) \ll 1$;
    \item \textbf{Mixed}: a dimension that, across the data, alternates between the active and passive
    conditions (i.e.\ its per-datapoint membership varies with $\rvx$).
\end{itemize}
We restate the mixed case as alternation across datapoints rather than as a set intersection, matching its
intended meaning. In practice the mixed class is hard to certify, since it requires partitioning the data
per dimension; \cite{Bonheme2021BeMA} therefore treat any dimension that is neither clearly active nor
clearly passive as mixed while the regime holds. Under strong regularisation even this can fail:
\cite{Clapham2023PosteriorCI} reports dimensions that satisfy none of the three conditions.

\par

Two features of this definition matter for us. First, the \emph{active} condition lives on the variance
representation ($\bar{\rvsigma}\ll1$), whereas the \emph{passive} condition is the one that constrains the
mean ($|\bar{\rvmu}|\ll1$, $\Var(\rvmu)\ll1$). A statistic of the mean alone can therefore hope to recover
the passive (collapsed) side but not the active side, which is exactly the side we want. Second, the whole
definition is anchored to the standard VAE's encoder statistics, so for other architectures a polarised
regime cannot be defined this way at all.
\cite{Bonheme2021BeMA} additionally observed that passive variables exhibit high correlation scores with
other variables in the \emph{mean} representation while being uncorrelated in sampled ones; we return to
this in Section~\ref{sec:emp_content}.

\subsection{The Active-Units Criterion}
\label{sec:au}

The most widely used collapse statistic predates all of the above. \cite{Burda2016IWAE} define the activity
of a latent dimension $u$ as
\begin{equation}
    A_u = \mathrm{Cov}_{\rvx}\big(\mathbb{E}_{u \sim q(u|\rvx)}[u]\big),
    \label{eqn:au}
\end{equation}
i.e.\ the variance, across the data, of the mean representation, and declare the dimension \emph{active}
if $A_u > 10^{-2}$. Two justifications are given: the distribution of $A_u$ on their trained models formed
two widely separated modes, and removing the inactive dimensions changed test log-likelihood by less than
$0.06$ nats. The metric has since become a standard reporting column in the posterior-collapse literature
\citep{He2019Lagging,Norouzi2020Exemplar,Razavi2019Delta,Menon2022Forget} and has been generalised per
stochastic layer for hierarchical models. \cite{zhang-etal-2022-effect} informally identify AU-inactive dimensions with the near-prior pathology
(variance near one, mean near zero), i.e. with Bonheme's passive condition, but no proof of that
correspondence exists in the literature. Two analytical treatments also bear on collapse. \cite{Sicks2021GLM} derive $A_u$ in closed form for a
generalised-linear $\beta$-VAE, and \cite{Li2026PhaseTransition} give a function-space stability analysis
of the collapse threshold, but both address when collapse should occur under Gaussian-decoder and
standard-prior assumptions. Here we instead use AU as a post-hoc collapse criterion.

Its fixed threshold has no calibration against the estimator, sample size, or latent scale. As
Section~\ref{sec:emp_agree} shows, this turns out to matter much less in practice than the arbitrariness
of $10^{-2}$ might suggest: in a polarised regime the collapsed and non-collapsed modes are so widely
separated that a broad range of thresholds gives the same classification.

\subsection{Statistical Quantities and their Estimation}
\label{sec:statistics}
\label{sec:approximation}

The collapse criteria considered here are functions of variance, entropy, KL divergence, or mutual
information. Figure~\ref{fig:bridge} summarises the relationships derived in
Section~\ref{sec:theory}, with the entropy of the mean as their common reference quantity.

\begin{figure}[ht]
    \centering
    \resizebox{0.95\linewidth}{!}{%
    \begin{tikzpicture}[
        font=\small,
        stat/.style={draw=black!70, rounded corners=2.5pt, fill=black!5,
                     align=center, inner sep=5pt, minimum height=9.5mm},
        hub/.style={stat, fill=blue!12, thick},
        holds/.style={-{Stealth[length=2.6mm]}, semithick, black!65},
        conv/.style={-{Stealth[length=2.6mm]}, semithick, black!65, dashed},
        lab/.style={font=\scriptsize\itshape, black!80, align=left, inner sep=1.5pt}
    ]
    \node[hub] (H) {$\mathrm{H}(\rvmu_i)$\\[-1pt]{\footnotesize entropy of the mean}};
    \node[stat, above=16mm of H] (KL)  {KL mean term\\[-1pt]$\mathbb{E}[\rvmu_i^2]$};
    \node[stat, right=17mm of H] (AU)  {AU / $\mathrm{Var}(\rvmu_i)$\\[-1pt]{\footnotesize(Burda et al.)}};
    \node[stat, left=17mm of H]  (BON) {Bonheme\\[-1pt]passive cond.};
    \node[stat, below=16mm of H] (MI)  {$\mathrm{MI}(\mX;\rvz_i)$\\[-1pt]{\footnotesize sampled-code rate}};

    \draw[holds] ([xshift=-1.6mm]KL.south)  -- ([xshift=-1.6mm]H.north);
    \draw[conv]  ([xshift= 1.6mm]H.north)   -- ([xshift= 1.6mm]KL.south);
    \draw[holds] ([yshift= 1.6mm]AU.west)   -- ([yshift= 1.6mm]H.east);
    \draw[conv]  ([yshift=-1.6mm]H.east)    -- ([yshift=-1.6mm]AU.west);
    \draw[holds] ([yshift= 1.6mm]BON.east)  -- ([yshift= 1.6mm]H.west);
    \draw[conv]  ([yshift=-1.6mm]H.west)    -- ([yshift=-1.6mm]BON.east);
    \draw[holds] ([xshift=-1.6mm]MI.north)  -- ([xshift=-1.6mm]H.south);
    \draw[conv]  ([xshift= 1.6mm]H.south)   -- ([xshift= 1.6mm]MI.north);

    \node[lab, anchor=north west] at ([xshift=3.5mm, yshift=-2.5mm]KL.south)
        {nested-family\\bound (Sec.~\ref{sec:ent_kld})};
    \node[lab, anchor=south, align=center] at ([yshift=5.8mm]$(AU.west)!0.5!(H.east)$)
    {entropy--variance bound\\(Sec.~\ref{sec:ent_kld})};
    \node[lab, anchor=south, align=center] at ([yshift=6.2mm]$(BON.east)!0.5!(H.west)$)
        {KL $\Leftrightarrow$ passive\\(Sec.~\ref{sec:ent_bon})};
    \node[lab, anchor=south west] at ([xshift=3.5mm, yshift=2.5mm]MI.north)
        {entropy-power sandwich\\(Prop.~\ref{prop:epi})};

    \begin{scope}[shift={([yshift=-7mm]MI.south)}]
        \draw[holds] (-3.1,0) -- (-2.3,0);
        \node[font=\scriptsize, anchor=west] at (-2.15,0) {always holds};
        \draw[conv]  (0.6,0)  -- (1.4,0);
        \node[font=\scriptsize, anchor=west] at (1.55,0) {conditional converse};
    \end{scope}
    \end{tikzpicture}}
    \caption{Map of the collapse statistics and the results connecting them, with the entropy of the mean
    $\mathrm{H}(\rvmu_i)$ as the hub. The mutual-information node is the finite rate of the noisy sampled
    code $\rvz_i$, not the degenerate population mutual information of a deterministic continuous mean.
    Solid arrows mark directions that hold under the stated model assumptions; dashed arrows mark converses
    requiring additional conditions, notably a small negentropy gap. Each edge is labelled with the section
    that derives it.}
    \label{fig:bridge}
\end{figure}

\par

\paragraph{Relation to the mutual-information / rate view.}
Characterising which latent dimensions are ``used'' through information theoretic quantities is not new.
Rate-based decompositions of the ELBO \citep{Hoffman2016ELBOSurgery,Alemi2018FixingAB} and
information-balancing objectives \citep{Zhao2019InfoVAE} analyse the aggregate rate $\mathrm{I}(\mX;\rvz)$
and its trade-off with distortion, and rate-derived MI estimates are used directly as collapse metrics
\citep{Menon2022Forget}. Our use is narrower and complementary: rather than an aggregate rate term, we use
a \emph{per-dimension}, prior-agnostic statistic $\mathrm{H}(\rvmu_i)$ defined directly on the mean
representation, which therefore applies unchanged to deterministic models, where no rate term exists.

\paragraph{Estimation.}
We distinguish Shannon entropy $\mathrm{H}(\cdot)$ for discrete variables from differential entropy
$\mathrm{h}(\cdot)$ for continuous variables in the theoretical analysis (Section~\ref{sec:theory}); in
empirical sections we write $\mathrm{H}(\cdot)$ generically. All latent and mean representations here are
continuous-valued, so the estimators approximate differential entropy, and all entropies are reported in
nats; we use the estimates for the \emph{relative ordering and separation} of dimensions, never as
calibrated absolute information content. We use the matrix-based R\'enyi $\alpha$-entropy estimator
\citep{renyi1961measures,Yu2018UnderstandingAW} at $\alpha=1.01$ (as $\alpha\to1$ R\'enyi entropy tends
to Shannon, here differential, entropy), which avoids explicit density estimation, supports joint
entropies between variables of different dimensionality, and behaves well empirically under appropriate
hyper-parameters \citep{Lee_2021}. Against histogram, $k$-nearest-neighbour (Kozachenko--Leonenko), and
Gaussian-mixture Monte-Carlo alternatives, compared on the synthetic control of
Section~\ref{sec:var_ent} (Figure~\ref{fig:spike}), it offered the best trade-off between cost and
stability, and we adopt it throughout.

\subsection{The Estimator Noise Floor}
\label{sec:floor}

Any entropy estimator returns a non-zero value even for an (almost) constant input, owing to finite $N$
and the estimator's settings; small estimated entropies are therefore meaningful only relative to this
floor. We measure the floor as a \emph{reference distribution}: at the exact operating point of an experiment,
the estimator is applied to $K$ independent draws of a control dimension that carries no information,
under two constructions: (i) a dimension frozen to a constant plus negligible numeric jitter, and (ii)
an i.i.d.\ noise dimension matched in scale to a collapsed unit. In all experiments below the floor is
built from $K=5{,}000$, each using $N=64$ samples
per batch and the Silverman bandwidth, with the noise scale set to the measured deep-collapse scale of the
trained models ($\sigma=0.0062$). It has mean $\bar{\mathrm{H}}_{\text{floor}}=0.0005$ nats, standard deviation
$s_{\mathrm{floor}}=1\times10^{-4}$ nats, and 95th percentile $q_{95}=0.0007$ nats. A dimension is \textbf{at
floor} when its estimated entropy lies below $q_{95}$ (one-sided from above, because
estimator noise is two-sided and a genuinely collapsed unit can land below the floor's mean). Verdicts of Section~\ref{sec:emp_gen} rest on unanimity across independent
runs rather than on any single dimension clearing $q_{95}$. We use the 95th percentile rather than the
observed maximum because the maximum of $K$ draws grows with $K$, which would make the floor depend on the
calibration budget, whereas a quantile converges as $K$ grows; the choice is not load-bearing, since the
reference distribution's maximum lies near $0.0009$ nats, so a max-based floor would shift every multiple
in Table~\ref{tab:h0report} by about a factor of $1.3$ and change no verdict. The floor is
\emph{not} the definition of collapse. Indeed, the detector of Section~\ref{sec:entropy_met} reads collapse off
each model's own fall-off, which dwarfs it (Section~\ref{sec:emp_agree}); its value is interpretive,
distinguishing a collapsed dimension at the absolute floor from one on a \emph{raised} floor, a
distinction Section~\ref{sec:emp_gen} turns on and which cannot be stated without a measured floor.

\section{Theoretical Analysis}
\label{sec:theory}

The established criteria above approach collapse through different statistics. For a standard VAE, the
per-dimension KL contribution is
\begin{equation}
\label{eqn:KL}
    D_{KL}(q(\rvz|\rvx)\,\|\,p(\rvz)) =
    \tfrac{1}{2}\big( \mathbb{E}[\rvmu_i^2] + \mathbb{E}[\rvsigma_i^2]
    - \mathbb{E}[\log \rvsigma_i^2] - 1\big),
\end{equation}
with a dimension deemed collapsed when its KL stays below a threshold $\epsilon$ for a sufficiently large
data fraction $\delta$ \citep{Lucas2019UnderstandingPC}. Together with the AU statistic
(Eq.~\ref{eqn:au}) and Bonheme's structural conditions, these appear to define collapse in different
ways. We now show that their mean-dependent components admit a common information-theoretic account,
summarised in Figure~\ref{fig:bridge}.

\subsection{A Ladder of Predictive Families: Entropy, Variance, and the KL}
\label{sec:ent_kld}

Three mean-side statistics are in play. The KL (Eq.~\ref{eqn:KL}) contains one term on the mean,
$\mathbb{E}[\rvmu_i^2]$, alongside two variance terms; the AU statistic is the variance of the mean,
$\Var(\rvmu_i)$; and our criterion is its entropy, $\mathrm{H}(\rvmu_i)$. These three are connected
by the $\mathcal{V}$-entropy of \cite{xu2020iclr-theory}: for a family $\mathcal{V}$ of probability
densities on $\mathbb{R}$,
\begin{equation}
    \mathrm{H}_{\mathcal{V}}(\rvmu_i) \;=\; \inf_{f\in\mathcal{V}}\ \mathbb{E}\big[-\log f(\rvmu_i)\big],
    \label{eqn:ventropy}
\end{equation}
the best expected log-loss any member of $\mathcal{V}$ attains on $\rvmu_i$'s marginal. In other words, it measures how
unpredictable $\rvmu_i$ is to a model who must pick a density from $\mathcal{V}$. A larger family
means a better model and a lower infimum, so nested families give ordered entropies.
 
\par
 
Three families, in increasing size, recover the statistics in play. A model with no choice at
all, the singleton $\mathcal{V}_0=\{\mathcal{N}(0,1)\}$ (the prior), pays the cross-entropy
$\tfrac12\log 2\pi + \tfrac12\,\mathbb{E}[\rvmu_i^2]$: up to constants, the KL mean term. A
model allowed to fit a mean and a variance, $\mathcal{V}_{\mathcal{G}}=\{\mathcal{N}(m,s^2)\}$,
does best at the matched moments and pays $\tfrac12\log\!\big(2\pi e\,\Var(\rvmu_i)\big)$: a
monotone function of the AU statistic (Eq.~\ref{eqn:au}). A model with a free choice of density,
$\mathcal{V}_{\mathrm{all}}$, pays, by Gibbs' inequality, the differential entropy
$\mathrm{h}(\rvmu_i)$ itself: our statistic. Since
$\mathcal{V}_0\subset\mathcal{V}_{\mathcal{G}}\subset\mathcal{V}_{\mathrm{all}}$, the three costs
are ordered:
\begin{equation}
    \tfrac12\log 2\pi + \tfrac12\,\mathbb{E}[\rvmu_i^2]
    \;\ge\; \tfrac12\log\!\big(2\pi e\,\Var(\rvmu_i)\big)
    \;\ge\; \mathrm{h}(\rvmu_i).
    \label{eqn:bound}
\end{equation}
The chain says the KL mean term, the AU statistic, and the entropy of the mean are one functional
evaluated under three family restrictions: each existing criterion asks how predictable $\rvmu_i$ is,
differing only in how much its model may fit; this is the map Figure~\ref{fig:bridge} draws. The
gap between the last two rungs is the negentropy of $\rvmu_i$, its KL divergence to the moment-matched
Gaussian.

\paragraph{The direction that always holds.}
The right-hand inequality of Eq.~\ref{eqn:bound} says a \emph{small variance forces a small entropy}.
This is the direction collapse detection needs for completeness: a passive dimension has
$\Var(\rvmu_i)\ll1$ (Bonheme's condition), so its entropy is correspondingly small. Thus an
AU-inactive dimension cannot retain a high marginal entropy merely through non-Gaussian shape. This
establishes the mean-variation half of the informal identification of AU-inactive dimensions with the
near-prior pathology \citep{zhang-etal-2022-effect}; Section~\ref{sec:ent_bon} supplies the prior-centring
and posterior-variance conditions enforced by KL minimisation.

\paragraph{The converse, and its condition.}
Reading collapse \emph{off} a low entropy is the converse, and it does \emph{not} hold in general.
The obstruction is exactly the negentropy gap: a low-entropy multimodal distribution can have large
variance. The converse is reliable when that gap is small, equivalently when $\rvmu_i$ is close to its
moment-matched Gaussian so that entropy and log-variance are in near-bijection. Crucially, concentration
alone does not imply this condition: scaling an arbitrarily non-Gaussian variable toward zero makes it
near-degenerate without changing its standardised shape or negentropy. We therefore treat small
negentropy as an empirical property to be checked, not a consequence of collapse. In the trained passive
dimensions studied here it is well supported: entropy and variance give nearly identical orderings and
classifications (Section~\ref{sec:emp_agree}), and Appendix~\ref{app:ivae_active} shows that the passive
distributions remain close to the Gaussian entropy--variance relation even where active iVAE dimensions
do not. A near-binary dimension (two narrow spikes at $\pm a$) remains the instructive counterexample:
it can have low entropy and substantial variance, so entropy alone cannot universally recover the
variance criterion.

\par

\paragraph{The rate on the sampled code.}
Everything above is a statement about $\rvmu_i$, while mutual information enters differently. If $\mX$ and $\rvmu_i=f(\mX)$ are treated as continuous
variables, $\mathrm{I}(\mX;\rvmu_i)$ is degenerate (generically infinite); the familiar identity
$\mathrm{I}(\mX;\rvmu_i)=\mathrm{H}(\rvmu_i)$ is valid only for a discrete empirical distribution
over datapoints and should not be read as a differential-entropy identity. The field's rate-based collapse
metrics instead use the noisy sampled code $\rvz_i$, whose posterior noise makes the information finite.
Under one further assumption, constant per-dimension noise, the mean-side entropy and variance quantities
bracket that rate:

\begin{proposition}[Entropy-power sandwich]
\label{prop:epi}
Let $\rvz_i = \rvmu_i + \sigma_i\varepsilon$ with $\varepsilon\sim\mathcal{N}(0,1)$ independent of $\mX$ and
$\sigma_i$ \emph{constant} across inputs, and let $N(\rvmu_i)=\tfrac{1}{2\pi e}e^{2\mathrm{h}(\rvmu_i)}$
denote the entropy power. Then
\begin{equation}
    \tfrac12\log\!\Big(1+\tfrac{N(\rvmu_i)}{\sigma_i^2}\Big)
    \;\le\; \mathrm{I}(\mX;\rvz_i)
    \;\le\; \tfrac12\log\!\Big(1+\tfrac{\Var(\rvmu_i)}{\sigma_i^2}\Big),
    \label{eqn:epi_sandwich}
\end{equation}
with equality on both legs iff $\rvmu_i$ is Gaussian.
\end{proposition}

\noindent The two legs are the max-entropy bound and the entropy-power inequality applied to the same
channel; Appendix~\ref{app:epi} defines the entropy power, records the two properties used, and gives the
derivation in full.

The two legs do not reverse the ordering of Eq.~\ref{eqn:bound}; they inherit it. The entropy-power
inequality $N(\rvmu_i)\le\Var(\rvmu_i)$ is exactly that equation's last rung rewritten, and
$x\mapsto\tfrac12\log(1+x/\sigma_i^2)$ is increasing, so the sandwich serves equivalently as the entropy and variance rungs. The rate is bracketed because entropy and variance give complementary bounds on the noisy channel. The upper bound says that a mean representation with little variance cannot transmit much information through posterior noise. The lower bound follows from the entropy-power inequality: whatever spread the mean does possess cannot be completely erased by adding independent Gaussian noise. The gap between the two reflects the non-Gaussianity of the mean representation. For collapse detection
only the upper leg is load-bearing, and it points the same way as Eq.~\ref{eqn:bound}: small
$\Var(\rvmu_i)$ forces a small rate.

\par

Gaussianity is not required for the collapse limit itself. Since
$0\le N(\rvmu_i)\le\Var(\rvmu_i)$, contraction of the mean representation
$\Var(\rvmu_i)\to0$ forces $N(\rvmu_i)\to0$ as well; both bounds in
Eq.~\ref{eqn:epi_sandwich} therefore tend to zero, and the sampled-code rate is squeezed to zero.
What Gaussianity controls is the \emph{width} of the sandwich away from that limit: a small negentropy
gap makes the entropy-power and variance bounds nearly coincide, while a strongly non-Gaussian mean can
separate them. Thus rate, entropy power, and variance share the same collapse implication without being
numerically interchangeable. One note regarding the constant-$\sigma_i$ assumption: when the encoder's
variance representation varies with the input, $\mathrm{h}(\rvz_i\mid\mX)$ is no longer a single
constant and the vanilla EPI decomposition used above breaks. This will be the case when dealing with
mixed variables.

\par

Finally, the KL can stay large through its variance terms even when the mean carries little
data-dependent variation. \cite{Takida2021PreventingPC} note a related effect: collapse can occur despite
non-negligible KL when the variance is fixed and only the mean is learned. Their analysis characterises
collapse through $\mathrm{I}(\mX;\hat{\mX})$, which involves the decoder and so mixes representation
collapse with decoder smoothness. We instead diagnose the mean representation directly through its
marginal entropy $\mathrm{h}(\rvmu_i)$, using Proposition~\ref{prop:epi} only when relating that
mean-side quantity to the finite rate of the noisy sampled code.

\subsection{KL Minimisation Enforces the Passive Condition}
\label{sec:ent_bon}

Section~\ref{sec:ent_kld} recovers the half of Bonheme's passive condition that lives on the mean,
$\Var(\rvmu_i)\ll1$, which is all a statistic of the mean can reach
(Section~\ref{sec:polarised_regime}). The variance half follows instead from the objective. Consider
minimising the KL (Eq.~\ref{eqn:KL}), the setting that produces collapse. The mean and variance
terms are minimised at $\bar{\rvmu}=0$ and $\bar{\rvsigma}^2=1$ respectively: the variance contribution
$f = \rvsigma_i^2 - \log\rvsigma_i^2 - 1$ vanishes iff $\rvsigma_i^2=1$ for all data, and the mean
contribution vanishes iff $\rvmu_i=0$. Together these are exactly Bonheme's passive condition. A
non-collapsed dimension is, conversely, one whose KL stays large, matching the entropy statistic under the
coupling of Section~\ref{sec:ent_kld}. What the variance representation could add for the \emph{mixed} case is deferred
to Appendix~\ref{app:mixed}.

\subsection{Where Entropy Beats Variance}
\label{sec:var_ent}

Thresholding the variance of the mean has been the field's collapse discriminator since
\cite{Burda2016IWAE} (Section~\ref{sec:au}). If variance already detects collapse, and is cheaper, what
does entropy add? We answer in two parts: the ladder of Section~\ref{sec:ent_kld} supplies the
justification the practice has lacked, and a constructed control identifies the one regime in which the
practice fails, leaving variance as the cheap default and entropy as the principled fallback.

\par

\begin{figure}[ht]
    \centering
    \includegraphics[width=0.7\linewidth]{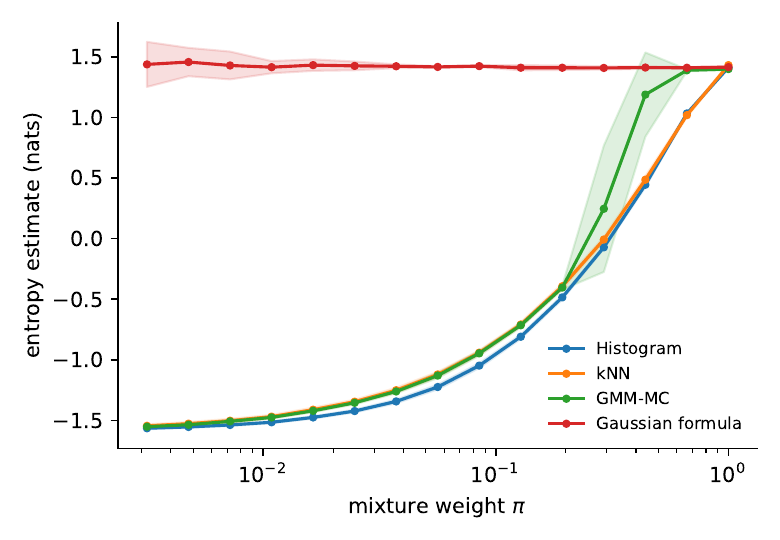}
    \caption{Continuous spike-and-slab control with $\Var(\rvz)$ held constant as the mixture weight $\pi$
    varies. Differential entropy of $\rvz$ rises monotonically from the collapsed ($\pi\!\to\!0$) to the
    informative ($\pi\!\to\!1$) extreme while variance is fixed, shown for several estimators (histogram,
    $k$-nearest-neighbours, Gaussian-mixture Monte-Carlo). Entropy separates the regimes that variance, by
    construction, cannot.}
    \label{fig:spike}
\end{figure}

\emph{Fixed second moments.} Entropy is sensitive to distributional
structure beyond the variance, so the two diverge when the second moment is held fixed. To isolate this, we
construct a continuous spike-and-slab variable
$\rvz \sim (1-\pi)\,\mathcal{N}(0,\epsilon^2) + \pi\,\mathcal{N}(0,\sigma^2)$, with $\epsilon=0.05$ and
$\sigma^2$ chosen so that $\Var(\rvz)$ is constant as $\pi$ varies. As $\pi$ moves from $0$ (collapsed) to
$1$ (informative), the variance is fixed by construction but the entropy varies monotonically between the
two extremes (Figure~\ref{fig:spike}): the negentropy gap of Section~\ref{sec:ent_kld} held open at fixed
variance (the same gap that sets the width of the sandwich in Proposition~\ref{prop:epi}). This is the regime (and, among those we study, the only one) in which entropy strictly out-resolves variance; no AU
threshold at any value can separate these cases. It is also entropy's \emph{only} genuine advantage here:
both are statistics of the mean alone, equally prior-agnostic, applying unchanged to deterministic models; prior-agnosticism separates this pair from the KL and encoder-statistic criteria, not entropy
from variance. On standard data their orderings match, so variance is the safe default and entropy the
principled fallback when second moments are pinned.

\section{Empirical Evaluation}
\label{sec:experiments}

We use the entropy of the mean to recover the polarised regime across architectures, quantifying the
identifiable-VAE raised floor against the calibrated floor, and then validate the unification of
Section~\ref{sec:theory} at scale, showing that the calibrated criterion and the frozen AU constant agree
on essentially all dimensions, and why. What the collapsed dimensions still carry is taken up in
Section~\ref{sec:emp_content}.

\subsection{Collapse Criterion and Thresholding}
\label{sec:entropy_met}

Operationally, we read collapse off the marginal entropy of each dimension. Let $\rvmu(\rvx)$ be the
encoder mean for input $\rvx \sim \mathcal{D}$ (for a deterministic model the latent code \emph{is}
the mean); then
\begin{equation}
\label{eq:entropy_metric}
    \text{collapsed}(i) \iff \mathrm{H}(\rvmu_i) \le \tau,
    \text{not collapsed}(i) \iff \mathrm{H}(\rvmu_i) > \tau,
\end{equation}
with $\tau$ placed inside the fall-off that separates the tight floor cluster from the informative
dimensions (the same bimodality \cite{Burda2016IWAE} observed for variance) by the automatic
construction of Appendix~\ref{app:tau}; when no fall-off exists, the detector reports \emph{no polarised
regime} rather than forcing a split. It is deliberately binary: it identifies the low-entropy cluster
associated with collapse and makes no attempt to separate active from mixed dimensions
(Appendix~\ref{app:mixed}). The reverse implication from low entropy to low variance is not assumed
universally; Section~\ref{sec:emp_agree} tests the small-negentropy condition under which the two criteria
coincide in the trained passive regime. Since neither entropy nor variance is reparameterisation-invariant
($\mathrm{h}(a\rmX)=\mathrm{h}(\rmX)+\log|a|$), ``small'' presupposes a mechanism that pins the
latent scale: a prior, a Lipschitz constraint, a norm penalty; every model we study carries one, and the
same presupposition attaches, unstated, to the AU statistic.

\subsection{Experimental Setup}
\label{sec:setup}

Each model class is evaluated in its canonical configuration, because the claim under test is generality
across model classes, not an architectural comparison between them: the LV-AE follows
\cite{qiuyi2024compressing} ($nz=128$, $32\times32$ inputs), including the Lipschitz-constrained decoder on
which its collapse mechanism depends; the iVAE follows the auxiliary-variable construction of
\cite{Khemakhem2019VariationalAA} with the conditional-prior parameterisation used by \cite{bonheme2023the}
($nz=30$, $64\times64$); the $\beta$-VAE uses the convolutional architecture of
\cite{Locatello2019ChallengingCA} ($nz=10$, $64\times64$); the L2-AE shares the standard convolutional
backbone ($nz=10$, $32\times32$) and serves as the non-collapsing control. All within-model quantities (the fall-off, the position of the passive cluster relative to the calibrated floor) are
self-referenced, and $\mathrm{H}_{\text{floor}}$ is calibrated at matched estimator settings
(Section~\ref{sec:floor}), so heterogeneous architectures do not enter any cross-model claim.

\paragraph{Model classes.}
\label{sec:models}
The $\beta$-VAE optimises Eq.~\ref{eqn:elbo}; a high $\beta$ can induce a polarised regime
\citep{rolinek2019VAEPCA} with disentangled active units
\citep{Bengio2013RepresentationLA,Locatello2019ChallengingCA}, and larger $\beta$ drives total posterior
collapse \citep{Bowman2016GeneratingSF}. iVAEs \citep{Khemakhem2019VariationalAA} address the
non-identifiability of the VAE latent space by conditioning on auxiliary information $\rvu$ (e.g.\ the
class label), optimising
\begin{equation}
   \mathcal{L}_{\text{iVAE}} = \mathbb{E}_{q_\phi(\rvz|\rvx, \rvu)} \left[ \log p_\theta(\rvx|\rvz) \right]
   - D_{\text{KL}} \left( q_\phi(\rvz|\rvx, \rvu) \,||\, p(\rvz|\rvu) \right),
    \label{eqn:ielbo}
\end{equation}
with a conditional prior $p(\rvz|\rvu)$; because this prior differs from the standard Gaussian, the
prior-based definition of collapse no longer applies as stated \citep{bonheme2023the}, and the
conditional prior writes the auxiliary variable into \emph{every} latent dimension, collapsed or not; this is arguably the mechanism behind Section~\ref{sec:emp_content}. LV-AEs \citep{qiuyi2024compressing} penalise the
volume of the latent space, the product of per-dimension variances, under a Lipschitz-constrained
decoder that blocks the trivial rescaling escape,
\begin{equation}
    \mathcal{L}_{\text{vol}} = \mathcal{L}^{\text{MSE}}(\rvx, \rvx')
    - \lambda \sqrt[m]{\textstyle\prod_i \left(\Var(\rvz_i) + \eta\right)},
    \label{eqn:lvol}
\end{equation}
with $\eta$ a small constant avoiding $\Var(\rvz_i)\approx0$, $\lambda$ the penalty strength, and $m$
the latent dimensionality: a polarised regime in a \emph{deterministic} model. The L2-AE penalises latent
magnitude,
\begin{equation}
    \mathcal{L}_{\mathrm{L2}}(\mathbf{x}, \mathbf{x}')
    = \| \mathbf{x} - \mathbf{x}' \|_2^2 + \lambda\, \| \mathbf{z} \|_2^2.
    \label{eq:l2ae}
\end{equation}
imposing no prior and no KL term; we expect a weak effect on the latent space and use it as the
non-collapsing control: a correct detector should find no polarised regime here.

\par

We train $\beta$-VAEs and iVAEs with $\beta\in\{1,2,4,8,16\}$, LV-AEs with
$\lambda\in\{3\mathrm{e}{-4},1\mathrm{e}{-3},3\mathrm{e}{-3},1\mathrm{e}{-2},3\mathrm{e}{-2}\}$ and
$\eta=1$, and L2-AEs over
$\lambda\in\{1,2,4,8,16\}$.
All models are optimised with Adam ($\beta_1=0.9$, $\beta_2=0.999$, $\epsilon=10^{-8}$) at learning rate
$10^{-4}$, with batch size $64$ for the $\beta$-VAE, iVAE, and L2-AE and $100$ for the LV-AE. The
$\beta$-VAE and iVAE are trained for $3\times10^{5}$ optimiser steps (three iVAE runs stop between
$2.55$ and $2.95\times10^{5}$); the LV-AE follows the $400$-epoch schedule of
\cite{qiuyi2024compressing} on which its volume-collapse mechanism is calibrated (one run stops at
$390$); and the L2-AE, which serves only as a non-collapsing control and is not compared quantitatively
with the others, is trained for $2\times10^{4}$ steps. Full per-model training configuration and a
reproducibility checklist are given in the supplementary material.
We use the benchmark datasets of \cite{Locatello2019ChallengingCA}: MNIST
\citep{LeCun1998GradientbasedLA}, smallNORB \citep{LeCun2004LearningMF}, and dSprites \citep{dsprites17},
with three seeds per configuration: $4$ model classes $\times$ $3$ datasets $\times$ $5$ hyperparameter
values $\times$ $3$ seeds $=180$ trained models, every one evaluated with the same estimator settings and
read against the same measured floor (Section~\ref{sec:floor}).

\subsection{The Polarised Regime Across Architectures}
\label{sec:emp_gen}
\label{sec:emp_coh}

\begin{figure*}[t]
    \centering
    \begin{subfigure}[b]{0.24\textwidth}\centering
        \includegraphics[width=\textwidth]{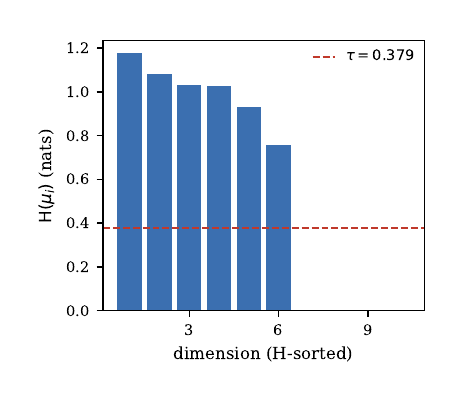}
        \caption{$\beta$-VAE, $\beta=2$}\label{fig:regime_bvae}\end{subfigure}\hfill
    \begin{subfigure}[b]{0.24\textwidth}\centering
        \includegraphics[width=\textwidth]{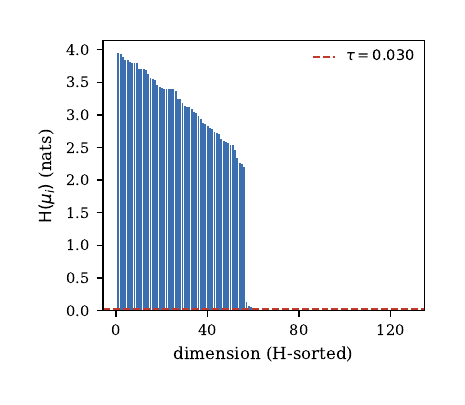}
        \caption{LV-AE, $\lambda=1\mathrm{e}{-3}$}\label{fig:regime_lvae}\end{subfigure}\hfill
    \begin{subfigure}[b]{0.24\textwidth}\centering
        \includegraphics[width=\textwidth]{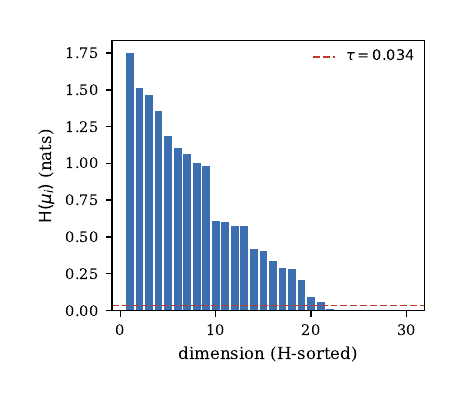}
        \caption{iVAE, $\beta=4$}\label{fig:regime_ivae}\end{subfigure}\hfill
    \begin{subfigure}[b]{0.24\textwidth}\centering
        \includegraphics[width=\textwidth]{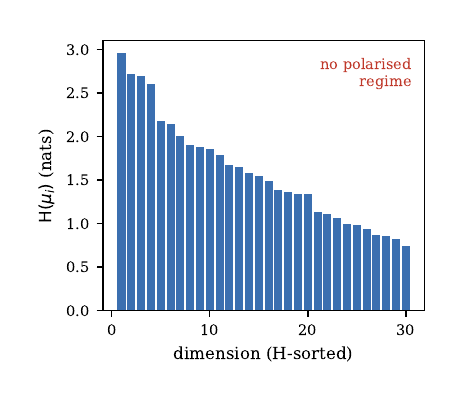}
        \caption{iVAE, $\beta=1$}\label{fig:regime_abstain}\end{subfigure}
    \caption{The polarised regime recovered by one criterion across architectures: sorted marginal
    entropies with the automatic threshold. (a) $\beta$-VAE (smallNORB); (b) the deterministic LV-AE
    (MNIST), with no prior or KL term; (c) iVAE (MNIST); (d) the same iVAE at $\beta=1$, where the detector
    abstains.}
    \label{fig:regimes}
\end{figure*}

\paragraph{Dynamics and the fall-off.}
Over training, entropy and variance follow the same trajectory for a typical active dimension, rising as
the encoder discovers structure and then plateauing, empirically matching the ordering of
Eq.~\ref{eqn:bound}. The matrix-based empirical estimate labelled
$\mathrm{MI}(\mathbf{X};\rvmu)$ follows the same trajectory as well, but we use it only as a finite-sample
consistency check: for the discrete empirical distribution over datapoints a deterministic map gives
$\mathrm{I}(\mX;\rvmu)=\mathrm{H}(\rvmu)$, whereas population mutual information for a continuous
deterministic encoder is degenerate (Section~\ref{sec:ent_kld}). The residual empirical
$\mathrm{H}(\rvmu\mid\mX)$ therefore reflects estimator bias/floor rather than a finite population
conditional entropy. Applying Eq.~\ref{eq:entropy_metric}, we
observe a clear collapsed/non-collapsed separation throughout training: non-collapsed dimensions rise to a high plateau while collapsing dimensions
fall rapidly toward the estimator floor. A common variant shows entropy rising then falling: a dimension
useful early in training but ultimately redundant and regularised away; we document these ``attempted
actives'' in Appendix~\ref{app:attempted}. After convergence the marginal entropy distribution shows the
fall-off shape the threshold construction targets: a tight floor cluster and well-separated informative
dimensions (Figure~\ref{fig:regimes}a). This is the same bimodality that \cite{Burda2016IWAE} reported
for their variance statistic.

\paragraph{LV-AEs.}
The same criterion recovers collapse in the deterministic LV-AE (Figure~\ref{fig:regimes}b), where no
prior or KL term exists to support the classical definitions. Small $\lambda$ leaves a dense spectrum of
informative dimensions (no polarised regime); larger $\lambda$ collapses everything; intermediate
$\lambda$ produces a clean split. Here non-collapsed dimensions reach entropies far above the
$\beta$-VAE's, since nothing pins the latent scale to a unit prior, which is why any \emph{fixed} constant is
in principle dependent on the model class, and why the per-model fall-off is the transferable object.

\paragraph{iVAEs: collapse to a (slightly) raised floor.}
iVAEs also exhibit the polarised structure (Figure~\ref{fig:regimes}c,d): large $\beta$ yields a graded
active spectrum above a tight floor cluster, small $\beta$ spreads entropy as a continuum with no floor
cluster (the detector reports no polarised regime). Our criterion required no adjustment to detect such a regime, in contrast to the prior-based definition, which had to be restated in terms of the
conditional prior \citep{bonheme2023the}.

What is distinctive about the iVAE is \emph{where} its floor cluster sits. A passive set, concretely, is a
tight low cluster of dimensions bunched just above the estimator floor and separated from the actives by a
large gap: in a representative iVAE (smallNORB, $\beta=2$), $19$ of the $30$ dimensions fall within
$[0.0016, 0.0085]$ nats (a median of about $6\times$ the floor), while the lowest active dimension sits
thirty-fold higher, at $0.24$ nats. Table~\ref{tab:h0report} reports, for every model/dataset cell, how
many of the fifteen runs split and where that cluster lies, summarising each cell by the \emph{median}
floor-cluster entropy over its splitting runs (the per-run value is heavy-tailed in $\beta$, so a mean
would track its largest run rather than a typical one). Against the measured floor ($q_{95}=$
$0.0007$ nats) the three identifiable-VAE families separate cleanly. The $\beta$-VAE's passive clusters
sit \emph{on} the floor (median $0.6$--$0.8\times$): its collapsed dimensions are indistinguishable from
information-free controls. The iVAE's sit robustly above it: a median of $5$--$7\times$ the floor, and,
decisively, \emph{every} one of its splitting runs lands at least $3\times$ above, on every dataset and
seed. The LV-AE is intermediate and strongly hyperparameter-dependent: its typical run sits near the floor
(median $0.5$--$4.3\times$) with a long upper tail (up to $36\times$ on a single run), so we do not read a
consistent raised floor into it. This \emph{raised floor} is small in absolute terms: a few thousandths of a
nat, irrelevant to classification (Section~\ref{sec:emp_agree} shows every reasonable threshold lands in
the same place). Its significance is interpretive, and only a measured floor makes it statable at all: the
iVAE alone retains, in every run, a systematic and model-attributable trace of structure in its collapsed
dimensions. Section~\ref{sec:emp_content} shows that this trace, invisible at marginal-statistic
amplitude, is most likely the auxiliary label.

\begin{table*}[t]
\centering
\caption{The polarised regime across 180 runs, read against the measured floor. \textbf{split}: runs (of
15 per cell) in which the detector finds a polarised regime; the remainder show a continuum or total
collapse. \textbf{passive $\tilde{\mathrm{H}}(\rvmu)$}: median floor-cluster entropy over the splitting
runs (nats). \textbf{$\times$ floor}: the same as a multiple of the floor's 95th percentile $q_{95}$ ($0.0007$ nats,
Section~\ref{sec:floor}), with $[\min,\max]$ over splitting runs; $\approx\!1$ means statistically at the
estimator floor. \textbf{verdict}: \emph{raised} = every splitting run clearly above the floor; \emph{at
floor} = not resolvable above the information-free control; \emph{variable} = straddles it. The
$\beta$-VAE/MNIST cell has no polarised regime because at $nz{=}10$ MNIST keeps all ten dimensions active;
LV-AE/MNIST carries a convergence caveat (9 of 15 runs still declining at the final checkpoint).}
\label{tab:h0report}
\begin{tabular}{llcccc}
\toprule
Model & Dataset & split & passive $\tilde{\mathrm{H}}(\rvmu)$ (nats) & $\times$ floor [min, max] & verdict \\
\midrule
iVAE & MNIST & 9/15 & $0.0036$ & $5.1\ [3.6, 6.6]$ & raised \\
iVAE & smallNORB & 15/15 & $0.0033$ & $4.7\ [3.1, 10.0]$ & raised \\
iVAE & dSprites & 15/15 & $0.0049$ & $7.0\ [3.9, 59.0]$ & raised \\
$\beta$-VAE & MNIST & 0/15 & \multicolumn{3}{c}{\emph{no polarised regime (all $nz{=}10$ dims active)}} \\
$\beta$-VAE & smallNORB & 14/15 & $0.0004$ & $0.6\ [0.3, 1.9]$ & at floor \\
$\beta$-VAE & dSprites & 8/15 & $0.0005$ & $0.8\ [0.6, 1.3]$ & at floor \\
LV-AE & MNIST & 11/15 & $0.0030$ & $4.3\ [0.0, 17.2]$ & variable \\
LV-AE & smallNORB & 14/15 & $0.0004$ & $0.5\ [0.1, 36.5]$ & variable \\
LV-AE & dSprites & 10/15 & $0.0013$ & $1.8\ [0.0, 4.9]$ & variable \\
L2-AE & MNIST & 0/15 & \multicolumn{3}{c}{\emph{no polarised regime}} \\
L2-AE & smallNORB & 0/15 & \multicolumn{3}{c}{\emph{no polarised regime}} \\
L2-AE & dSprites & 0/15 & \multicolumn{3}{c}{\emph{no polarised regime}} \\
\bottomrule
\end{tabular}
\end{table*}

\par

The raised floor invites comparison with the no-collapse result of \cite{wang2021posterior}, but theirs
concerns a \emph{reversible-encoder} notion of identifiability while our iVAEs are of the
auxiliary-variable kind, so we make no claim to verify or reconcile the two. As a sanity check we trained
their reversible-encoder construction and observed the same polarised regime, although we report this only as a preliminary observation.

\subsection{Empirical Support for Unification}
\label{sec:emp_agree}

Section~\ref{sec:theory} identifies the negentropy gap as the obstruction to reversing the
entropy--variance bound. We therefore test whether that obstruction is negligible in the trained passive
regime; dimension-by-dimension, it is. Of the 180 runs, 96 exhibit a polarised
regime (iVAE $39/45$, $\beta$-VAE $22/45$, LV-AE $35/45$, L2-AE $0/45$; the non-collapsing control, correctly, never splits). Across the splitting runs' $\approx\!5{,}900$ dimensions, the calibrated entropy
criterion and Burda's frozen $A_u>10^{-2}$ rule disagree on fewer than $1\%$ of dimensions
($52$ dimensions in total; roughly three-quarters of runs have \emph{zero} disagreement, and the residual
disagreements concentrate in the LV-AE, at most $5$ of $128$ dimensions in any run, where nothing ties
$\Var(\rvmu)$ to the unit scale the constant presumes). Even on the single worst-disagreeing run of each
model class, the two thresholds still land in the same fall-off gap; the disagreements are edge cases at
the margin, never a wholesale relabelling.

\par

\begin{figure}[ht]
    \centering
    \includegraphics[width=0.7\linewidth]{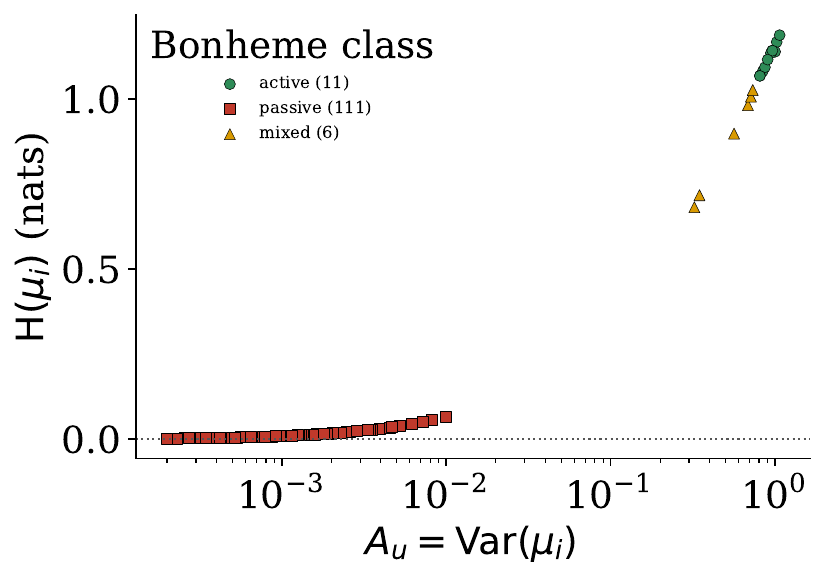}
    \caption{Per-dimension differential entropy $\mathrm{H}(\mu_i)$ versus variance $A_u=\mathrm{Var}(\mu_i)$
    for a high-capacity $\beta$-VAE ($nz=128$, $\beta=4$, MNIST), each dimension coloured by its full
    \cite{Bonheme2021BeMA} class. The dotted line marks the estimator floor.}
    \label{fig:entvar}
\end{figure}

The agreement is visible within a single model before it is counted at scale.
Figure~\ref{fig:entvar} shows per-dimension differential entropy against variance for a high-capacity
$\beta$-VAE ($nz=128$, MNIST), with dimensions coloured by their full Bonheme class. The passive
dimensions form a tight low-entropy, low-variance cluster, the active dimensions occupy the opposite
extreme, and the few mixed dimensions bridge the two. The passive cluster therefore lies in the
small-negentropy regime identified in Section~\ref{sec:ent_kld}, where the entropy--variance bound is
effectively reversible for classification. That MNIST exhibits a polarised regime here is itself
informative: at the canonical $nz=10$ it uses every dimension, whereas with surplus capacity
($nz=128$) all but approximately $11$ dimensions collapse. The regime is therefore induced by surplus
capacity rather than being an intrinsic property of the dataset.

\par

Why, then, does the fixed $A_u>10^{-2}$ rule work so reliably? Because the fall-off is enormous. Across
every splitting run, the floor cluster and the informative dimensions are separated by
$10^{3}$--$10^{4}$ standard deviations of the measured estimator floor, so almost any threshold placed
inside the gap yields the same classification. The empirically tight entropy--variance relationship in
the passive regime carries this robustness over to the AU statistic. The value $10^{-2}$ is therefore
not special; it simply happens to lie inside a very wide separation. The rule becomes questionable only
when that separation disappears or when the latent scale is not pinned, as in the small residual
disagreements observed for the LV-AE. The calibrated criterion additionally abstains when no polarised
regime is present, rather than forcing a split.

\section{What Collapsed Dimensions Still Carry}
\label{sec:emp_content}

Everything so far (AU, KL, encoder statistics, and our own calibrated entropy criterion) is a
\emph{marginal} statistic: a function of one dimension's distribution, measuring the amplitude of its
variation. This section shows that amplitude is a systematically incomplete description of a collapsed
dimension, in two steps: first the correlation structure of the passive block, then its decodable content.

\subsection{Passive Correlation}
\label{sec:shadow}

\cite{Bonheme2021BeMA} reported that passive variables show high correlation scores with other variables
in the mean representation (while being uncorrelated in sampled ones). We reproduce this and identify its
structure. Classifying dimensions by Bonheme's own conditions and computing correlations across the mean
representation: active dimensions are mutually decorrelated (mean $|\rho|\approx0.03$--$0.04$ for the
$\beta$-VAE and iVAE) while passive dimensions correlate both with the actives \emph{and with each other}
at a similar, much higher level ($0.17$--$0.23$): the signature of a shared source rather than of
pairwise coupling among the passives. Two candidate sources exist, the active subspace itself and, in the
iVAE, the shared low-amplitude copy of $\rvu$ (Section~\ref{sec:beyond_label}); the pairwise structure
alone does not separate them. Whatever its source, the shared signal is \emph{faint}, because the dimensions carrying it are
collapsed: their variance sits at the deep-collapse scale of Section~\ref{sec:floor}, orders of magnitude
below the actives (Figure~\ref{fig:entvar}). High correlation on the mean representation
therefore does not imply information content: correlation is invariant to per-dimension rescaling
($\mathrm{corr}(a\rmX,\rmY)=\operatorname{sign}(a)\,\mathrm{corr}(\rmX,\rmY)$ for $a\neq0$), so
it is blind to what entropy measures, amplitude, and is in that sense the first of the
scale-equivariant probes that drive the rest of this section.

\subsection{Downstream: The Collapsed Block Decodes the Label}
\label{sec:downstream}
\begin{figure}[ht]
    \centering
    \includegraphics[width=.85\linewidth]{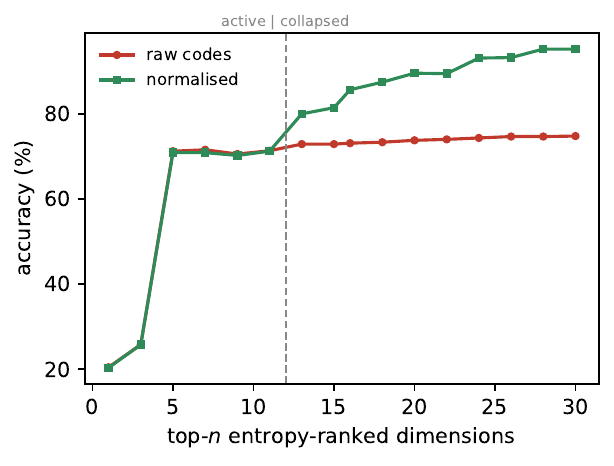}
    \caption{Accuracy against the number of top-entropy-ranked dimensions for a representative iVAE on
    smallNORB (raw vs normalised codes; vertical line at the calibrated active/collapsed boundary).
    Accuracy continues to climb well past the boundary: the collapsed dimensions carry decodable label
    structure.}
    \label{fig:downstream}
\end{figure}

If the raised floor were merely a shadow, removing what the actives explain would empty it. We test
content directly: logistic classifiers predict the ground-truth label from (i) the active dimensions
alone, (ii) actives plus the collapsed dimensions, at raw scale and after per-dimension standardisation,
and (iii) actives plus matched-scale i.i.d.\ noise dimensions in place of the collapsed ones (the
capacity control). Table~\ref{tab:downstream} reports the results pooled over splitting runs. For the
$\beta$-VAE and LV-AE the pattern refines \cite{Burda2016IWAE}'s ablation: raw-scale gains are nil,
normalised gains are modest ($4$--$11$ points). The iVAE is
qualitatively different. On smallNORB, active-only accuracy is $63.8\%$; adding the collapsed dimensions
yields $96.7\%$ ($+33$ points; random control $61.2\%$); on dSprites, $43.1\%$ rises to $80.1\%$
($+37$). The gain does not require rescaling: at \emph{raw} scale the collapsed block already lifts
smallNORB accuracy from $62.6\%$ to $75.2\%$, and the collapsed dimensions \emph{alone}, with no actives at all, decode the label at $67.7\%$ raw ($87.5\%$ normalised) against a $20\%$ nominal ($21.2\%$
empirical majority-class) chance rate. And the signal
is not borrowed from the actives: regressing the active means out of the collapsed block before decoding
leaves the result essentially unchanged, by $-0.12$ points on smallNORB ($96.35\%\to96.22\%$) and
$+0.15$ points on dSprites ($80.13\%\to80.28\%$), both smaller than the variation across train/test splits.
By contrast the $\beta$-VAE's and LV-AE's collapsed blocks alone decode at chance at raw scale ($23.7\%$ and $25.9\%$
on smallNORB). In particular for identifiable models, collapse does not erase content: the conditional prior $p(\rvz|\rvu)$
writes the label into every dimension, and the collapsed ones retain it at an amplitude no marginal
statistic registers; quantitatively, the between-class variance $\Var_{\rvu}(\mathbb{E}[\rvmu_i|\rvu])$
of the iVAE's passive dimensions is an order of magnitude larger than the $\beta$-VAE's
($3.7\times10^{-5}$ vs $3.7\times10^{-6}$ on smallNORB) at comparable total passive variance.
Figure~\ref{fig:downstream} is the picture: accuracy climbs well past the calibrated
active/collapsed boundary once codes are normalised.

\paragraph{Why standardisation isolates linearly decodable content.}
Per-dimension standardisation does not create linear signal. For any invertible diagonal rescaling $\mD$, a linear decision rule $\rvw^\top\rvz+b$ on the raw code has an equivalent rule $(\mD^{-1}\rvw)^\top\mD\rvz+b$ on the rescaled code. An exact, unregularised linear-probe family is therefore closed under rescaling: its attainable predictions are unchanged, as the usable ($\mathcal{V}$-)information framework states more generally \citep{xu2020iclr-theory}. The normalised columns of Table~\ref{tab:downstream} accordingly ask the scale-free question of this section: whether the collapsed dimensions retain linearly decodable label structure once their amplitude is factored out. The raw columns ask a different, practical question: whether the same structure is accessible to a regularised, finitely optimised probe at its native scale. Low-amplitude features require correspondingly large weights, which the $\ell_2$ penalty discourages; a raw--normalised gap therefore reflects reduced accessibility at the collapsed scale, not information
created by standardisation. Weakening the penalty confirms this directly: sweeping $C$ upward on the
iVAE closes the gap from $20.7$ to $0.4$ points on smallNORB and from $21.5$ to $0.1$ points on dSprites,
with raw accuracy rising to meet the normalised value that barely moves. Marginal entropy remains sensitive to precisely this scale suppression, since
$\mathrm{h}(a\mX)=\mathrm{h}(\mX)+\log|a|$: shrinking a coordinate lowers its entropy even when its linear predictive structure is preserved up to rescaling. The two instruments thus answer different questions. Marginal collapse statistics quantify variation at the representation's native amplitude; normalised probes test the task-relevant linear structure that remains after amplitude is removed.

\begin{table*}[t]
\centering
\caption{Downstream classification accuracy (\%), pooled over the splitting
$(\text{hyperparameter},\text{seed})$ runs of each model/dataset. All codes are standardised per dimension,
so accuracy reflects \emph{content}, not amplitude. ``Active only'' uses the non-collapsed dimensions;
``+\,collapsed'' adds the collapsed dimensions; ``+\,random'' instead adds matched-scale i.i.d.\ Gaussian
noise, an amplitude-matched control that contributes no label information. Probes are multinomial logistic
regressions (\textsc{lbfgs}, $\ell_2$, $C=1$, tol $10^{-4}$, max\_iter $500$, intercept fitted).}
\label{tab:downstream}
\begin{tabular}{llcccc}
\toprule
Model & Dataset & Active only & +\,collapsed (norm.) & +\,random \\
\midrule
iVAE & MNIST & $97.4$ & $98.5$ & $96.7$ \\
iVAE & smallNORB & $63.8$ & $96.7$ & $61.2$ \\
iVAE & dSprites & $43.1$ & $80.1$ & $42.0$ \\
$\beta$-VAE & smallNORB & $54.1$ & $60.5$ & $53.4$ \\
$\beta$-VAE & dSprites & $44.1$ & $50.4$ & $44.3$ \\
LV-AE & MNIST & $78.4$ & $86.7$ & $74.4$ \\
LV-AE & smallNORB & $68.4$ & $79.0$ & $65.0$ \\
LV-AE & dSprites & $43.1$ & $47.5$ & $42.7$ \\
\bottomrule
\end{tabular}
\end{table*}

\subsection{Coalition Analysis}
\label{sec:coalition}
\begin{figure*}[t]
    \centering
    \includegraphics[width=0.65\textwidth]{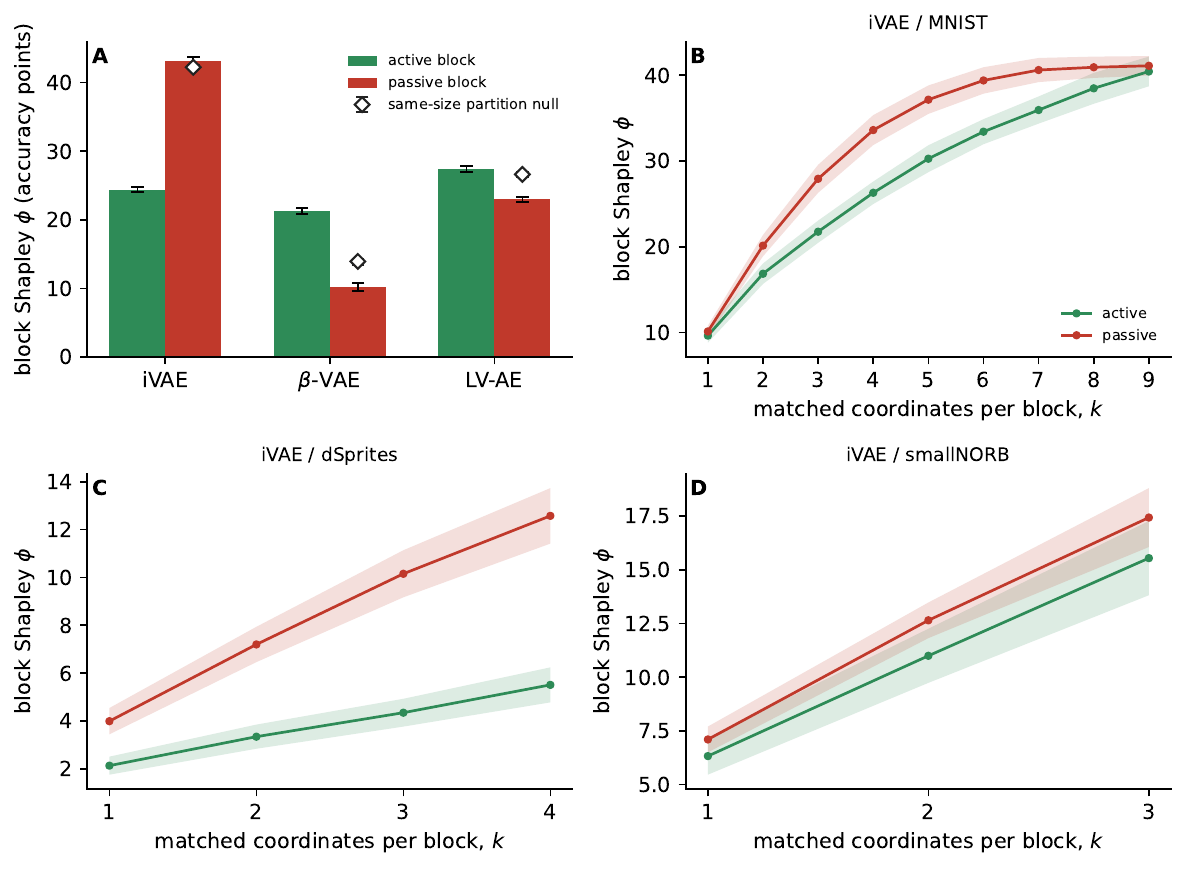}
    \caption{Cardinality-controlled Shapley analysis, with every coordinate standardised to unit variance.
    \textbf{A}: whole-block attribution averaged across the $20$ independent splits; error bars are the
    2.5th--97.5th percentiles of the split-level model-class means, and open diamonds mark the passive
    attribution expected from block size alone ($150$ random partitions per checkpoint and split).
    \textbf{B--D}: iVAE matched-cardinality curves from $50$ nested subset draws per checkpoint, each point
    comparing $k$ active with $k$ passive coordinates; bands are standard errors across checkpoint means,
    and each panel stops at the largest $k$ every checkpoint in that dataset supports.}
    \label{fig:shapley}
\end{figure*}

The downstream gain is a property of the collapsed \emph{block}; how much of the joint decoding that block
is actually responsible for, once amplitude is removed as a confound, is answered cleanly by its Shapley
value. Standardising every dimension to unit variance, so that the collapsed dimensions are ``scaled up'' to the actives, we attribute the joint label accuracy to the active and collapsed blocks by their exact
two-block Shapley value, averaged over the clean two-sided splits and checked against a label-shuffle
control that drives both values to zero (Figure~\ref{fig:shapley}). Across the split resampling, once amplitude is equalised, the
collapsed block accounts for a \emph{substantial} share of the decodable label in every model class:
$42\%$ for the $\beta$-VAE, $49\%$ for the LV-AE, and $67\%$ for the iVAE. This recovers, and quantifies,
the observation that motivated this study: collapse suppresses the \emph{amplitude} of a dimension's
variation far more than its decodable \emph{content}, and a large fraction of that content is retrievable
in \emph{every} model once the scale is corrected.

\par

The block shares themselves, however, must not be read as per-coordinate quality: they are confounded by
block \emph{cardinality}. Across checkpoints the passive share tracks the fraction of coordinates called
passive (Pearson $r=0.74$ overall, $0.74$--$0.86$ within each model class), so a large share partly
records how many dimensions collapsed rather than what each carries. We therefore read every observed
block against a same-size random-partition null: coordinates randomly reassigned to two blocks of the
original cardinalities, with the decoder and Shapley attribution rerun. We repeat the full comparison over
$20$ independent $70/30$ splits, with $150$ partitions per checkpoint and split. On the primary
accuracy-point scale, the iVAE passive block sits at its null (excess $+0.91$ points; split-level 95\% range
$[+0.62,+1.17]$; paired Wilcoxon on each checkpoint's 20-split mean, $p=0.253$), while the $\beta$-VAE's
and LV-AE's passives fall below theirs ($-3.74$ points, $p=0.025$; $-3.64$ points,
$p=7.1\times10^{-8}$). Read per coordinate, an iVAE passive dimension therefore contributes at least at
parity with an average coordinate of the same representation, actives included, whereas the other
classes' whole passive blocks contribute less than their coordinate count predicts.

A fixed-cardinality probe sharpens this result: decoding from three randomly chosen actives against three
randomly chosen passives, again with $150$ draws on every split, gives an iVAE passive-minus-active
difference of $+4.60$ Shapley accuracy points (split-level 95\% range $[+4.09,+5.00]$;
$p=3.9\times10^{-4}$). The $\beta$-VAE trend is negative ($-2.87$, $p=0.198$) and the LV-AE trend is
smaller and marginal ($+1.47$, $p=0.171$).
because passive coordinates are highly redundant with respect to the label, being distinct low-amplitude projections of shared structure (cf.\ Section~\ref{sec:beyond_label}), so small subsets look strong while
the full block adds little beyond its size. The non-trivial fact is therefore not that the collapsed \emph{block}'s share is large, which is mostly counting, but that its individually near-dead coordinates decode at parity with live ones once
amplitude is equalised, which the next subsection shows is genuine content and not an echo of the label.
The matched-cardinality curves make the redundancy visible (Figure~\ref{fig:shapley}B--D): on MNIST the
passive advantage is largest at small $k$ and has almost vanished by $k=9$, whereas on dSprites the passive
curve remains more than twice the active curve through the range shared by all $15$ checkpoints
(fixed-count difference $+6.66$ points, Holm-adjusted $p=1.8\times10^{-4}$; MNIST $+6.75$, Holm
$p=0.016$; smallNORB $+1.26$, Holm $p=0.389$). The dSprites cell
is the same one with the largest downstream gain in Table~\ref{tab:downstream}; because it was selected
from the same eight model--dataset cells, we treat it as the strongest exploratory interaction rather than
a confirmed effect. Dataset-level estimates and the Holm correction for the three iVAE fixed-count tests
are reported in Appendix~\ref{app:shapley}.

\par

Checkpoints share crossed seeds and hyperparameters, and the model classes differ in capacity and
architecture at once; both caveats are taken up in the limitations, and where a between-class comparison
is load-bearing, the per-dimension label density of the collapsed block, we match capacity explicitly
(Table~\ref{tab:nonu}).

\subsection{Beyond the Label}
\label{sec:beyond_label}
The downstream result concerns the auxiliary label $\rvu$, which is an \emph{input} to the iVAE encoder
$q(\rvz\mid\rvx,\rvu)$: decoding $\rvu$ from the collapsed dimensions could be dismissed as reading back the
answer the encoder was handed. Two controls show the retained content is genuine. \emph{First}, the
collapsed dimensions decode generative factors \emph{statistically independent} of $\rvu$, which a copy of
$\rvu$ cannot produce (Table~\ref{tab:nonu}): on dSprites ($\rvu={}$shape) the collapsed block lifts scale
decoding by $+10$ points and position by $+6$ over the active dimensions alone; on smallNORB
($\rvu={}$category) it lifts azimuth by $+4$ and elevation by $+3$. The count- and scale-matched noise
control carries none of this, sitting at or slightly below the active-only baseline in every row, so the
gain is content rather than capacity padding. \emph{Second}, and correcting an over-strong reading
of the raw-scale result, this retention does not require identifiability; \emph{surplus capacity} suffices: a $\beta$-VAE given the same over-provisioning ($nz=128$ on dSprites, $118$ collapsed
dimensions, and \emph{no} auxiliary variable) also decodes scale ($+7$), orientation ($+8$), and position
($+18$--$22$) from its collapsed block, and reads shape, its would-be label, at raw scale from those
dimensions ($53.7\%$ against $33.3\%$ chance). Collapse suppresses amplitude, not content, in \emph{any}
representation with more latent dimensions than the data require; the canonical $nz=10$ $\beta$-VAE shows
nothing here only because it retains too few passive dimensions.

\par

What \emph{is} specific to the identifiable model is narrower: the label is present in far more of the
collapsed dimensions, and more strongly in the best of them. Decoding shape from each collapsed dimension
\emph{individually} on dSprites, at matched raw scale and on held-out data, $94\%$ of the iVAE's
collapsed coordinates beat chance against $34\%$ of the capacity-matched $\beta$-VAE's, fewer than the
half a signal-free population would place above chance by sampling alone, and the strongest
single iVAE coordinate reaches $56.3\%$ against the $\beta$-VAE's $39.3\%$ (chance $33.3\%$) despite the
$\beta$-VAE offering $118$ coordinates to select from rather than the iVAE's twenty-odd. The label is
therefore widespread rather than confined to a favoured few, but the distribution is not flat: the iVAE's
median coordinate reaches $37.5\%$, so a strong upper tail sits above a broad base of weak-but-real
signal. The natural reading, which we offer as a hypothesis and do not establish, is that
$p(\rvz\mid\rvu)$ writes a dense low-amplitude copy of $\rvu$ into the dimensions the decoder declines
to recruit, each retaining a distinct projection, consistent with the collapsed block's independence from
the actives (Section~\ref{sec:downstream}: decoding survives regressing the active means out). We do
not characterise what each dimension retains, or whether it can be put to use; those we leave open.

\begin{table*}[t]
\centering
\caption{Collapse retains content \emph{independent} of the label: decode accuracy (\%) of generative
factors statistically independent of the conditioning variable $\rvu$, from the iVAE collapsed block,
amplitude-equalised. \textbf{active only} uses the non-collapsed dimensions;
\textbf{+random} is the count- and scale-matched noise control (active dimensions plus the same number of
matched-scale noise columns), which falls at or slightly below the active-only baseline in every row, so
lifts quoted in the text are measured against \textbf{active only}. The capacity-matched $\beta$-VAE
control and the per-dimension density comparison are reported in the text.}
\label{tab:nonu}
\begin{tabular}{lcccc}
\toprule
Dataset: factor ($\perp\rvu$) & chance & active only & +\,random & +\,collapsed \\
\midrule
dSprites: scale & $16.7$ & $41.0$ & $40.3$ & $51.2$ \\
dSprites: position & $3.1$ & $17.1$ & $14.2$ & $23.0$ \\
smallNORB: azimuth & $5.6$ & $14.7$ & $12.9$ & $18.5$ \\
smallNORB: elevation & $11.1$ & $16.1$ & $15.5$ & $19.4$ \\
\bottomrule
\end{tabular}
\end{table*}

\paragraph{Mean versus sampled representations.}
The content result is a statement about the \emph{mean} representation, and essentially so: a passive
dimension's posterior variance is $\approx\!1$ while its class-conditional mean separation is of order
$10^{-2}$, so in the sampled representation $\rvz=\rvmu+\rvsigma\odot\rvepsilon$ the label signal is
destroyed by the dimension's own posterior noise: decoding from sampled codes drops the passive block to
chance in every case while barely touching the active block (per-dataset numbers in
Appendix~\ref{app:meansampled}). This closes the loop with \cite{Bonheme2021BeMA}, who found passive
correlations present in the mean but absent in sampled codes, and explains how the disentanglement
literature's convention of evaluating on sampled codes \citep{Locatello2019ChallengingCA} never saw this
structure: it lives exactly where those evaluations do not look.

\section{Discussion and Limitations}
\label{sec:discussion}

\paragraph{Detector scope and scale.}
By design, the criterion is a binary collapse detector: it does not separate active from mixed
dimensions, both of which can have high marginal entropy of the mean. We regard the mixed case as a
separate problem, with the variance-entropy Bernoulli signal of Appendix~\ref{app:mixed} a plausible but
not-yet-validated route to it. Like AU, the entropy criterion is not reparameterisation-invariant: it
presupposes a mechanism that pins the latent scale (a prior, a Lipschitz constraint, or a norm penalty),
which every model studied here carries. The converse from low entropy to low variance additionally
requires a small negentropy gap; concentration alone does not guarantee it
(Section~\ref{sec:ent_kld}). In the trained passive regimes studied here that condition is empirically
well satisfied (Section~\ref{sec:emp_agree}), but it is not part of the definition of collapse.

\paragraph{Estimator dependence.}
The raised-floor reading depends on the measured estimator floor
(Section~\ref{sec:floor}). The collapse verdicts themselves are threshold-robust
(Section~\ref{sec:emp_agree}), but ``at floor'' versus ``raised'' is relative to
an estimator- and setting-dependent reference. The LV-AE/MNIST entry of
Table~\ref{tab:h0report} is further weakened by its convergence caveat and should
be read provisionally. More generally, the entropy estimates are used to compare
the relative amplitude and separation of dimensions, not as calibrated measures
of their absolute information content; Section~\ref{sec:emp_content} shows directly
that dimensions at very low marginal entropy can nevertheless retain decodable
structure.

\paragraph{Empirical scope.}
The empirical study is limited to image data and convolutional autoencoders; whether the same picture
holds for sequential \citep{Bowman2016GeneratingSF} or diffusion models is open. The L2-AE control is
also trained on a shorter schedule than the other classes (Section~\ref{sec:setup}); we rely on it only to
confirm that a weakly regularised autoencoder produces no fall-off, which is confirmed in all $45$ runs, and
draw no quantitative comparison from it. Low-entropy informative codes are therefore established here
within a deliberately bounded empirical setting, not claimed as a universal property of every collapsed
representation.

\paragraph{Between-model comparisons.}
The between-model-class comparisons of Section~\ref{sec:emp_content} (Shapley shares and raised-floor
magnitudes) vary latent capacity and architecture together. The random-partition null removes the
cardinality confound within each checkpoint, and split resampling removes dependence on a single
train/test partition, but neither removes capacity and architecture differences between classes; the
paired tests remain diagnostics over crossed (hyperparameter, seed) checkpoints rather than confirmatory
inference. The one between-class comparison on which we rely, the per-dimension label density of the
collapsed block, matches capacity explicitly but rests on a single $nz{=}128$ $\beta$-VAE checkpoint
(Section~\ref{sec:beyond_label}). A systematic capacity-controlled cross-model suite remains future work.

\section{Conclusion}
\label{sec:conclusion}

We connected the field's disconnected collapse statistics in a common account centred on the entropy of
the mean representation. The KL mean term, AU variance, and entropy are the same predictive functional
under nested family restrictions, while Proposition~\ref{prop:epi} carries the entropy-power and variance
quantities to the finite mutual-information rate of the noisy sampled code. The reverse
entropy-to-variance implication is not universal: its obstruction is the negentropy gap. Empirically that
gap is negligible for the trained passive dimensions, and across 180 models the calibrated entropy
criterion and the frozen AU constant label fewer than $1\%$ of dimensions differently. The familiar
$10^{-2}$ threshold works here not because that number is special, but because the polarised-regime
fall-off leaves orders of magnitude of slack.

\par

That agreement also sharpens what a collapse verdict does not say. These criteria are sensitive to the
native-scale strength of a dimension's variation; they do not certify the absence of scale-free decodable
structure. Once amplitude is equalised, collapsed dimensions carry decodable label content in every model
class and generative factors independent of the label. A capacity-matched control shows that this retention
does not require identifiability: surplus latent capacity suffices. What is specific to identifiable VAEs
is that their conditional prior imprints the auxiliary label into collapsed dimensions at high
per-dimension density, with nearly all such coordinates decoding it individually where a capacity-matched
$\beta$-VAE's sit at chance.

\par

The deterministic LV-AE licenses a broader reading: geometric volume regularisation alone, with no prior
and no KL term, induces the same fall-off signature. Posterior collapse is therefore one instance of a
broader information-collapse phenomenon for which a prior-agnostic, calibrated mean-side language is
useful, but whose boundary must be stated carefully: deciding that a dimension lies in the collapsed
low-amplitude cluster is threshold-robust in the regimes studied here; deciding what that dimension still
contains requires conditional, scale-invariant probes. Future work should validate a combined
mean/variance-entropy classifier for the mixed class, characterise content retention as a function of
regularisation strength, test sequential and diffusion-based generative models, and further formalise the
predictive-family ladder within the usable-information framework.

\section*{Declaration of generative AI and AI-assisted technologies in the writing process}

During the preparation of this work, the author(s) used a large language model--based assistant (Anthropic
Claude) to help draft and edit prose and to help develop the analysis and figure-generation scripts. After using this tool, the author(s) reviewed and edited all resulting text, independently checked every reported quantity against the underlying data and code, and take full responsibility for the content of the publication.

\section*{Declaration of competing interest}

The authors declare that they have no known competing financial interests or personal relationships that
could have appeared to influence the work reported in this paper.

\section*{Funding}

This work was supported by the Engineering and Physical Sciences Research Council (EPSRC) through a Doctoral
Training Partnership.

\section*{Disclaimer}

Where authors are identified as personnel of the International Agency for Research on Cancer/World Health
Organization, the authors alone are responsible for the views expressed in this article and they do not
necessarily represent the decisions, policy or views of the International Agency for Research on
Cancer/World Health Organization.

\section*{Data availability}

The datasets used in this study are publicly available from their original sources.

\bibliographystyle{cas-model2-names}
\bibliography{references}

\appendix

\section{Entropy Power and the Proof of Proposition~\ref{prop:epi}}
\label{app:epi}

\paragraph{Entropy power.}
For a continuous random variable $X$ on $\mathbb{R}$ with differential entropy $\mathrm{h}(X)$, the
\emph{entropy power} is
\begin{equation}
    N(X) \;=\; \frac{1}{2\pi e}\,e^{2\mathrm{h}(X)}.
    \label{eqn:entpower}
\end{equation}
It is best read as a variance. A Gaussian $\mathcal{N}(0,s^2)$ has
$\mathrm{h}=\tfrac12\log(2\pi e s^2)$, so solving $\mathrm{h}(X)=\tfrac12\log(2\pi e s^2)$ for $s^2$
returns exactly Eq.~\ref{eqn:entpower}: $N(X)$ is the variance of the Gaussian that is as uncertain as
$X$. This makes it the natural currency for comparing a non-Gaussian variable with the Gaussian noise it
is added to, which is what the proposition needs. Two standard properties are used.

\emph{(i) Entropy power never exceeds variance.} Among all variables of a given variance the Gaussian
maximises differential entropy, $\mathrm{h}(X)\le\tfrac12\log(2\pi e\Var(X))$. Exponentiating,
\begin{equation}
    N(X)\;\le\;\Var(X),\qquad\text{with equality iff }X\text{ is Gaussian.}
    \label{eqn:npvar}
\end{equation}
The ratio $\Var(X)/N(X)$ is $e^{2D}$, where $D$ is the negentropy of $X$: its KL divergence to the
moment-matched Gaussian, and the same gap that separates the last two rungs of Eq.~\ref{eqn:bound}.

\emph{(ii) The entropy-power inequality (EPI).} For independent $X$ and $Y$,
\begin{equation}
    N(X+Y)\;\ge\;N(X)+N(Y),
    \label{eqn:epi}
\end{equation}
with equality iff both are Gaussian \citep{shannon1948mathematical,STAM1959101}. Entropy powers of independent
summands add at least as fast as variances do.

\paragraph{Proof of Proposition~\ref{prop:epi}.}
Write $\rvz_i=\rvmu_i+\sigma_i\varepsilon$ with $\varepsilon\sim\mathcal{N}(0,1)$ independent of $\mX$ and
$\sigma_i$ constant across inputs. Conditioning on $\mX$ fixes $\rvmu_i$, leaving only the Gaussian noise,
so $\mathrm{h}(\rvz_i\mid\mX)=\tfrac12\log(2\pi e\sigma_i^2)$ and
\begin{equation}
    \mathrm{I}(\mX;\rvz_i)\;=\;\mathrm{h}(\rvz_i)-\tfrac12\log(2\pi e\sigma_i^2)
    \;=\;\tfrac12\log\frac{N(\rvz_i)}{\sigma_i^2},
    \label{eqn:rate_identity}
\end{equation}
the second equality by Eq.~\ref{eqn:entpower}. The constant-$\sigma_i$ assumption makes $\rvz_i$ an additive-noise channel whose Gaussian noise is
independent of $\rvmu_i$, and gives the fixed conditional entropy
$\mathrm{h}(\rvz_i\mid\mX)=\tfrac12\log(2\pi e\sigma_i^2)$. If $\sigma_i$ varies with the input,
neither simplification holds in this form, so the EPI argument below does not apply directly. Both legs now follow from bounding $N(\rvz_i)$.

For the \emph{upper} leg, apply Eq.~\ref{eqn:npvar} to $\rvz_i$ and use the independence of $\rvmu_i$ and
$\varepsilon$ to write $\Var(\rvz_i)=\Var(\rvmu_i)+\sigma_i^2$:
\begin{equation*}
    N(\rvz_i)\;\le\;\Var(\rvmu_i)+\sigma_i^2 .
\end{equation*}
Substituting into Eq.~\ref{eqn:rate_identity} gives
$\mathrm{I}(\mX;\rvz_i)\le\tfrac12\log\big(1+\Var(\rvmu_i)/\sigma_i^2\big)$.

For the \emph{lower} leg, apply the EPI (Eq.~\ref{eqn:epi}) to the independent sum, noting that
$N(\sigma_i\varepsilon)=\sigma_i^2$ because $\sigma_i\varepsilon$ is Gaussian:
\begin{equation*}
    N(\rvz_i)\;\ge\;N(\rvmu_i)+\sigma_i^2 .
\end{equation*}
Substituting into Eq.~\ref{eqn:rate_identity} gives
$\mathrm{I}(\mX;\rvz_i)\ge\tfrac12\log\big(1+N(\rvmu_i)/\sigma_i^2\big)$, completing
Eq.~\ref{eqn:epi_sandwich}.

Equality on the upper leg requires equality in Eq.~\ref{eqn:npvar} for $\rvz_i$, i.e.\ $\rvz_i$ Gaussian,
which by Cram\'er's theorem forces $\rvmu_i$ Gaussian. Equality on the lower leg requires equality in the
EPI, i.e.\ both summands Gaussian; $\sigma_i\varepsilon$ already is, so again $\rvmu_i$ must be. Both legs
are therefore tight exactly when $\rvmu_i$ is Gaussian. $\square$

\paragraph{What the sandwich is worth.}
Subtracting the legs, their separation is
$\tfrac12\log\big[(\sigma_i^2+\Var(\rvmu_i))/(\sigma_i^2+N(\rvmu_i))\big]$, which by
Eq.~\ref{eqn:npvar} is the negentropy gap of $\rvmu_i$ damped by the posterior-noise scale: the sandwich
is tight where $\rvmu_i$ is near-Gaussian and widens as its entropy power falls below its variance.
No Gaussian assumption is needed to obtain the collapse limit. Eq.~\ref{eqn:npvar} gives
$0\le N(\rvmu_i)\le\Var(\rvmu_i)$, so if $\Var(\rvmu_i)\to0$ then
$N(\rvmu_i)\to0$ by squeezing. Both bounds in Eq.~\ref{eqn:epi_sandwich} consequently tend to zero,
forcing $\mathrm{I}(\mX;\rvz_i)\to0$. Near-Gaussianity controls only how closely the two bounds track one
another away from that limit; it is not implied by near-degeneracy. The bound also inherits rather than
reverses the ordering of Eq.~\ref{eqn:bound}: Eq.~\ref{eqn:npvar} is that equation's last rung rewritten,
and $x\mapsto\tfrac12\log(1+x/\sigma_i^2)$ is increasing, so the two legs are the entropy-power and
variance rungs carried through one monotone map.

\section{Placing the Threshold $\tau$}
\label{app:tau}

In a polarised regime the per-dimension entropies show a characteristic shape: a \emph{tight cluster} of
floor-level dimensions, separated from the informative dimensions above by a \emph{fall-off}, a step
whose size dwarfs the floor cluster's internal spacing. \cite{Burda2016IWAE} observed exactly this
bimodality for their variance statistic and justified their constant by it; the gap, not the constant, is
the transferable object. We place $\tau$ by locating the floor directly: sorting the entropies in
ascending order, $\tau$ is set at the first (lowest) gap that is at least ten times the median
nearest-neighbour spacing of the dimensions below it, subject to the floor being a genuine cluster (at
least three dimensions, with total extent smaller than the gap) and the gap itself exceeding $100$ floor
standard deviations, i.e.\ far above anything estimator noise can produce. This ``smallest tight bottom
cluster with a cliff above it'' construction deliberately models the collapsed/informative boundary as a
\emph{fall-off} rather than a \emph{taper}: a graded active spectrum that tapers smoothly towards the
floor never drags $\tau$ upward, because the search stops at the first genuine cliff. The scale
presupposition of Section~\ref{sec:entropy_met} is sharpest here: a wholly unconstrained deterministic
autoencoder could hold an informative dimension at arbitrarily small scale (the decoder's first layer
compensating), which would read as collapsed; detecting collapse there would require a per-dimension
normalisation convention that we do not develop. When no such cliff exists (the entropies form a continuum, or every dimension is at the floor), the construction returns no
$\tau$ and the detector reports \emph{no polarised regime} rather than forcing a split. We note the
correct reading of that outcome: it is not that activity is undecidable, merely that there is no bimodal
collapse structure; dimensions of substantial variance are still active in the ordinary sense.

\section{The Mixed Case}
\label{app:mixed}

For completeness we note what the variance representation could add, without making it part of the
detector. One might try to take the entropy of the \emph{variance} representation, but this does not
separate active from passive: in both cases the variance is tightly concentrated (near $0$ and near $1$
respectively), hence low-entropy. Where it \emph{does} help is the mixed case. A mixed dimension's variance
alternates between $\approx0$ (active) and $\approx1$ (passive) across the data; in the idealised limit it
is Bernoulli with $\Pr[\text{passive}] = p$, giving
\begin{equation}
    \mathrm{H}(\rvsigma_i) = -(q \ln q + p \ln p), \qquad q = 1-p,
\end{equation}
maximised at $p=\tfrac12$, so the entropy of the variance assigns mixed dimensions strictly higher entropy
than either polarised extreme. Combined with the entropy of the mean this could in principle recover all
three classes. We have not validated such a combined classifier and do not rely on it; we flag it as the
natural route to the mixed case (Section~\ref{sec:discussion}).

\par

The KL offers a second route. For mixed dimensions, examine the pointwise KL: low for passive inputs, high for active inputs. A dimension
is mixed when some inputs give low KL and others high. This presupposes a polarised regime and breaks down
when the regime fails. Bonheme's criteria can \emph{detect} that failure (a dimension satisfying none of
the conditions) but then no longer support variable selection. The entropy detector remains defined, since per-dimension information is meaningful even without a polarised regime, and, consistent with its stated
scope, it continues to answer only the collapse question.

\section{The iVAE's Non-Gaussian Active Dimensions}
\label{app:ivae_active}

\begin{figure}[ht]
    \centering
    \includegraphics[width=\linewidth]{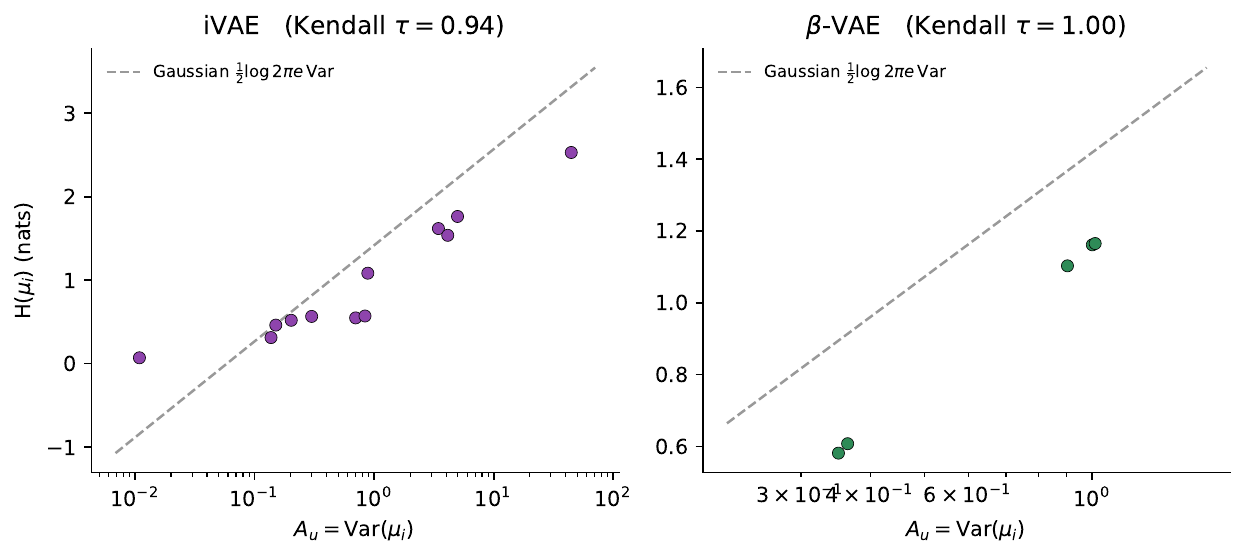}
    \caption{Per-dimension differential entropy against variance for the \emph{active} dimensions of a
    representative iVAE (left) and $\beta$-VAE (right), each the run whose rank agreement equals its
    model's median. The dashed line is the Gaussian relation
    $\mathrm{H}=\frac12\log 2\pi e\,\mathrm{Var}$.}
    \label{fig:monotonicity}
\end{figure}

Being a conditional model, the iVAE has no reason
to produce Gaussian means, and among its \emph{active} dimensions it does not: the $\beta$-VAE tracks the
Gaussian entropy--variance relation almost exactly (Figure~\ref{fig:monotonicity}, right; Kendall
$\tau=1.00$) while the iVAE departs from it (left; $\tau=0.94$; median $|\text{excess kurtosis}|$ $2.2$
versus $0.4$), so a higher-variance dimension can carry lower entropy: a naturally occurring instance of
the fixed-second-moment regime of Section~\ref{sec:var_ent}. It has no bearing on collapse detection: the
\emph{passive} dimensions remain close to the Gaussian entropy--variance relation in every model
($|\text{excess kurtosis}|$ $0.4$--$0.7$), providing empirical support for the small-negentropy condition
used to interpret the converse in Section~\ref{sec:ent_kld}. We record the active-side departure as a
curiosity, not a load-bearing result.

\section{Attempted Actives}
\label{app:attempted}
\label{sec:novel}

Training dynamics reveal ``attempted actives'': dimensions whose entropy rises over the first few dozen
epochs and then falls to the estimator floor well before convergence. These are transiently useful but ultimately suppressed by
regularisation pressure, a dynamic phenomenon that static definitions of collapse miss. This supports the
hypothesis of \cite{Bonheme2021BeMA} that dimensions are learned in order of reconstruction relevance, and
is consistent with collapse reflecting competitive allocation of representational capacity. We offer one
mechanism as a hypothesis for future work. The collapse thresholds of \cite{Sicks2021GLM} and
\cite{Li2026PhaseTransition} are computed against the \emph{residual} data variance, the variation not yet captured by the other dimensions, and that residual shrinks as competing dimensions improve. The boundary
is therefore not static: a dimension that sits above it early in training can be overtaken by it later and
pruned, which is exactly the attempted-active trajectory. A natural follow-up, given
Section~\ref{sec:emp_content}, is whether an attempted active in an identifiable model retains its content after being pruned, i.e.\ whether the label trace in the collapsed dimensions is the residue of early
learning or is written in by the conditional prior throughout.

\section{Split-Resampled Shapley Robustness}
\label{app:shapley}

Table~\ref{tab:shapley_split} separates uncertainty from the random coordinate partitions from uncertainty
in the downstream train/test split. Every eligible checkpoint is evaluated on $20$ independently seeded
$70/30$ splits; within each split the same-size partition null and fixed-$k=3$ comparison each use $150$
draws. The split interval is the empirical 2.5th--97.5th percentile of the model/dataset mean across those
$20$ splits. The Wilcoxon test instead uses one value per checkpoint after averaging that checkpoint over
all splits, preserving the trained checkpoint as the inferential unit. The three iVAE dataset-level
fixed-count tests form one family and receive Holm correction; dataset-level whole-block effects are shown
descriptively without additional post-hoc tests.

\begin{table*}[ht]
\centering
\caption{Split-resampled cardinality controls, in Shapley accuracy points. $\Delta\phi_{P,\mathrm{null}}$
is observed passive attribution minus the same-size random-partition expectation;
$\phi_P^{(3)}-\phi_A^{(3)}$ compares three passive with three active coordinates. Brackets give the
split-level 95\% range. Class-level rows are the main-text aggregates; Holm adjustment applies to the three
iVAE dataset fixed-count tests only.}
\label{tab:shapley_split}
\begin{tabular}{llrccccl}
\toprule
Model & Dataset & $n$ & $\Delta\phi_{P,\mathrm{null}}$ [split 95\%] & $p$ & $\phi_P^{(3)}-\phi_A^{(3)}$ [split 95\%] & $p$ & Holm $p$ \\
\midrule
iVAE & all datasets & 39 & $0.91\;[0.62, 1.17]$ & 0.253 & $4.60\;[4.09, 5.00]$ & 0.00039 & -- \\
$\beta$-VAE & all datasets & 22 & $-3.74\;[-4.12, -3.33]$ & 0.025 & $-2.87\;[-3.45, -2.07]$ & 0.198 & -- \\
LV-AE & all datasets & 33 & $-3.64\;[-3.85, -3.29]$ & 7.1e-08 & $1.47\;[1.18, 1.82]$ & 0.171 & -- \\
iVAE & MNIST & 9 & $-3.64\;[-3.96, -3.34]$ & -- & $6.75\;[6.03, 7.59]$ & 0.008 & 0.016 \\
iVAE & dSprites & 15 & $6.13\;[5.35, 6.82]$ & -- & $6.66\;[5.68, 7.76]$ & 6.1e-05 & 0.00018 \\
iVAE & smallNORB & 15 & $-1.59\;[-2.31, -1.06]$ & -- & $1.26\;[-0.13, 2.29]$ & 0.389 & 0.389 \\
\bottomrule
\end{tabular}

\end{table*}

\section{Mean versus Sampled Representations: Per-Dataset Numbers}
\label{app:meansampled}

Decoding from sampled codes (averaged over draws of $\rvepsilon$), the passive-block accuracy collapses to chance in every case: iVAE
smallNORB $87.5\%\!\to\!20.1\%$ (empirical chance $21.2\%$), MNIST $85.0\%\!\to\!10.4\%$ (chance
$11.6\%$), dSprites
$74.9\%\!\to\!33.3\%$ (chance $34.4\%$); $\beta$-VAE smallNORB $37.3\%\!\to\!19.7\%$. Meanwhile the
active-block accuracy, whose dimensions have $\sigma^2\!\approx\!0$, is barely touched (iVAE smallNORB
$62.3\%\!\to\!56.1\%$, MNIST $96.9\%\!\to\!95.6\%$). The smallNORB active baseline differs from
Table~\ref{tab:downstream}'s $63.8\%$ because the two analyses draw independent subsamples and splits.
The difference is not the sample size: over $20$ harmonised splits the pooled means at $N=2{,}500$ and
$N=3{,}000$ are $62.76\%$ and $62.80\%$, a gap of $0.04$ points, while the split ranges span
$61.98$--$63.79\%$ and $62.03$--$63.79\%$. The discrepancy is split jitter, and
Table~\ref{tab:downstream}'s value sits at the upper end of that range. The passive-only means are likewise stable ($87.63\%$ and $87.70\%$).

\end{document}